%% file: paper.tex
\documentclass[twoside,11pt]{article}
\usepackage{jair, theapa, rawfonts}

\pdfoutput=1

\usepackage{float}
\usepackage{physics}
\usepackage{amsmath}
\usepackage{mathtools} %multilined
\usepackage{siunitx} % scientific notation
\usepackage{multirow} % table
\usepackage{pgf}
\usepackage{fancyvrb} % BVerbatim
\usepackage{cleveref} % Cref: Figure(s), Section(s), Equation(s), Table(s)
\usepackage{caption}
\usepackage{subcaption} % subfigure
\usepackage[linesnumbered,ruled,vlined]{algorithm2e}
\usepackage[acronym,nomain]{glossaries}

\usepackage[utf8x]{inputenc}
\usepackage[T1]{fontenc} % for using unicode in bibtex
\usepackage{lmodern}

\usepackage[mode=buildnew]{standalone}
\usepackage{tikz}
\usetikzlibrary{shapes, arrows, snakes, backgrounds, positioning, calc}
\usetikzlibrary{shadows}

\makeatletter
\pgfdeclareshape{document}{
\inheritsavedanchors[from=rectangle]
\inheritanchorborder[from=rectangle]
\inheritanchor[from=rectangle]{center}
\inheritanchor[from=rectangle]{north}
\inheritanchor[from=rectangle]{south}
\inheritanchor[from=rectangle]{west}
\inheritanchor[from=rectangle]{east}
\backgroundpath{
\southwest \pgf@xa=\pgf@x \pgf@ya=\pgf@y
\northeast \pgf@xb=\pgf@x \pgf@yb=\pgf@y
\pgf@xc=\pgf@xb \advance\pgf@xc by-0pt
\pgf@yc=\pgf@yb \advance\pgf@yc by-0pt
\pgfpathmoveto{\pgfpoint{\pgf@xa}{\pgf@ya}}
\pgfpathlineto{\pgfpoint{\pgf@xa}{\pgf@yb}}
\pgfpathlineto{\pgfpoint{\pgf@xc}{\pgf@yb}}
\pgfpathlineto{\pgfpoint{\pgf@xb}{\pgf@yc}}
\pgfpathlineto{\pgfpoint{\pgf@xb}{\pgf@ya}}
\pgfpathcurveto{\pgfpoint{30pt}{-10pt}}{\pgfpoint{15pt}{-30pt}}{\pgfpoint{-3.5pt}{\pgf@ya}}
\pgfpathclose}}
\makeatother

\ShortHeadings{Improving QA Performance Using Knowledge Distillation and Active Learning}
{Boreshban, Mirbostani, Ghassem-Sani, Mirroshandel, \& Amiriparian}
\firstpageno{1}

\begin{document}

\title{Improving Question Answering Performance \\
Using Knowledge Distillation and Active Learning}

\author{\name Yasaman Boreshban \email boreshban@ce.sharif.edu \\
	\addr Computer Engineering Department, \\
	Sharif University of Technology, 
	Tehran, Iran \\
	\AND
	\name Seyed Morteza Mirbostani \email m.mirbostani@msc.guilan.ac.ir \\
	\addr Department of Computer Engineering, \\
	University of Guilan,
	Rasht, Iran \\
	\AND
	\name Gholamreza Ghassem-Sani \email sani@sharif.edu \\
	\addr Computer Engineering Department, \\
	Sharif University of Technology,
	Tehran, Iran \\
	\AND
	\name Seyed Abolghasem Mirroshandel \email mirroshandel@guilan.ac.ir \\
	\addr Department of Computer Engineering, \\
	University of Guilan,
	Rasht, Iran \\
	\AND
	\name Shahin Amiriparian \email shahin.amiriparian@uni-a.de \\
	\addr Chair of Embedded Intelligence for Health Care \& Wellbeing, \\
	University of Augsburg, 
	Augsburg, Germany
}

% For research notes, remove the comment character in the line below.
% \researchnote

\maketitle

% Acronyms
\newacronym{abs_qa}{QA}{question answering}
\newacronym{abs_kd}{KD}{knowledge distillation}
\newacronym{abs_al}{AL}{active learning}
\newacronym{qa}{QA}{question answering}
\newacronym{kd}{KD}{knowledge distillation}
\newacronym{skd}{SKD}{self-knowledge distillation}
\newacronym{al}{AL}{active learning}
\newacronym{nlp}{NLP}{natural language processing}
\newacronym{dnn}{DNN}{deep neural network}
\newacronym{rnn}{RNN}{recurrent neural network}
\newacronym{bert}{BERT}{Bidirectional Encoder Representations from Transformer}
\newacronym{cnn}{CNN}{convolutional neural network}
\newacronym{lstm}{LSTM}{long short-term memory}
\newacronym{mse}{MSE}{mean squared error}
\newacronym{use}{USE}{universal sentence encoder}
\newacronym{ner}{NER}{named entity recognition}
\newacronym{squad}{SQuAD}{Stanford Question Answering Dataset}
\newacronym{ir}{IR}{information retrieval}
\newacronym{rc}{RC}{reading comprehension}
\newacronym{kl}{KL}{Kullback-Leibler}
\newacronym{ce}{CE}{cross-entropy}

\begin{abstract}
	Contemporary \gls{abs_qa} systems, including transformer-based architectures, suffer from increasing computational and model complexity which render them inefficient for real-world applications with limited resources. Further, training or even fine-tuning such models requires a vast amount of labeled data which is often not available for the task at hand. In this manuscript, we conduct a comprehensive analysis of the mentioned challenges and introduce suitable countermeasures. We propose a novel \gls{abs_kd} approach to reduce the parameter and model complexity of a pre-trained BERT system and utilize multiple \gls{abs_al} strategies for immense reduction in annotation efforts. In particular, we demonstrate that our model achieves the performance of a 6-layer TinyBERT and DistilBERT, whilst using only 2\% of their total parameters. Finally, by the integration of our \gls{abs_al} approaches into the BERT framework, we show that state-of-the-art results on the SQuAD dataset can be achieved when we only use 20\% of the training data.
\end{abstract}

\section{Introduction}
\label{introduction}

The development of \gls{qa} systems is a relatively new challenge in the field of \gls{nlp} \cite{KOLOMIYETS20115412}. The ultimate goal of creating such systems is to enable machines to comprehend text as well as, or even better than, human beings \cite{zhang2019machine}. Extensive progress has been made in this area over the last few years. In \gls{qa} models, context paragraphs and their corresponding questions are represented as a series of tokens \cite{yu2018qanet}. The objective of a \gls{qa} system is to predict the correct span within a paragraph in which the answer to a given question resides. It is often the case that an attention mechanism is also used to keep the dependency relations between questions and paragraphs. Furthermore, two probability values are computed for each token, which represents the likelihood of the token being the start and end of an answer span. For each query, the system identifies the span with the highest probability value, as the answer to the query.

With the insurgence of interest in \glspl{dnn}, recent \gls{qa} models have achieved excellent results. On some corpora, they have even reached an accuracy level higher than humans. Nevertheless, these achievements have been made possible with the cost of building very large and expensive \gls{nlp} models. Despite all the progress made, there are still several remaining challenges and issues that need to be addressed. For instance, these models often suffer from high complexity and low robustness issues. Moreover, they normally require a massive amount of labeled data for training. These models usually have too many parameters, leading to a considerable training time. In addition, they are subject to extensive resource consumption for performant operation and reasonable inference time, which makes them unfit for real-world applications running on devices with limited resources such as mobile and embedded devices \cite{cheng2020survey}. Highly effective deep learning-based approaches can immensely enhance the performance of distributed systems, embedded devices, and FPGAs. The use of machine learning technology in virtual and augmented reality on hardware such as smart wearable devices has brought distinct accomplishments in terms of features and capabilities. However, due to the excessive computational complexity imposed by this technology, its implementation on most portable devices is challenging and bounded by their hardware limitations. Accordingly, to address this issue, different model compression techniques have been introduced as a practical solution, which has absorbed a lot of attention over the recent years.

Current compression techniques can be divided into four general groups of parameter pruning and quantization, low-rank factorization, transferred/compact convolutional filters, and \gls{kd} \cite{oguntola2018slimnets}. It has been suggested that among these methods, using \gls{kd} can result in a more significant improvement in terms of accuracy and performance. Accordingly, we have decided to study the impact of \gls{kd} on the \gls{qa} task.

Another concerning issue entangled with \glspl{dnn} is the robustness deficiency. Although employing \glspl{dnn} in \gls{nlp} models has led to impressive results on multiple downstream tasks, these models are not robust enough and are extremely vulnerable to adversarial examples. For \gls{qa} tasks, it has been demonstrated that an intentional perturbation of a paragraph through including adversarial sentences confuses even the best available \gls{qa} models, causing a severe reduction of their accuracy. This vulnerability against adversarial examples also makes these models unsuitable for real-world scenarios. Consequently, numerous studies addressing this issue have been conducted to increase the robustness of the proposed models \cite{jia2017adversarial}.

Recent accomplishments in \gls{dnn} have been heavily dependent on the use of large training datasets; conversely, \glspl{dnn} are inefficient when trained on small datasets; however, the number of available annotated corpora is inadequate, and manual annotation is a costly procedure. Moreover, for some languages, the required amount of annotated datasets is unavailable. In recent years, there has been a limited number of studies conducted on unsupervised, semi-supervised, and \gls{al} for \gls{qa} systems. In this study, we introduce a novel combination of a parameter reduction technique and \gls{al} for \gls{qa} systems. We show that the results of this combination are comparable to that of state-of-the-art models for this task.

For parameter reduction, we utilize \gls{kd} to transfer the knowledge of a large (complex) model to a condensed neural network. In other words, we train a small model in such a way that its accuracy would be close to that of the initial large model. In this study, we have used a pre-trained model as our initial model and transferred its knowledge to a small \gls{qa} model. It has been demonstrated that employing \gls{kd} significantly improves the robustness and generalization of the models \cite{7546524}. In this paper, we have specifically investigated the impact of \gls{kd} on the robustness of \gls{qa} systems. We also utilize \gls{al} to minimize the cost of data labeling. To the best of our knowledge, \gls{al} has not so far been applied to the task of \gls{qa}. Since data annotation is an expensive task, we can employ \gls{al} strategies to obtain reasonable results with a small training dataset. Generally, the primary goal of \gls{al} is to reach high accuracy with low-cost data labeling \cite{Fu2013}. During the \gls{al} process, we use several strategies to select informative unlabeled data samples with the ability to transfer more information to the model. Hence, we are able to drastically reduce the required number of samples and their labeling costs for training the model. By combining \gls{kd} and \gls{al} methods, we build a model with a significantly reduced number of parameters and required labeled samples. The resultant model is capable of achieving comparable results to that of state-of-the-art models.

The structure of this paper is as follows: we define the theoretical background of \gls{qa} systems in Section 2 and introduce our related works in Section 3. We describe our proposed approaches in detail in Section 4. We give a brief description of the datasets used in this study and present our experimental results in Section 5. Finally, Section 6 includes our conclusions and future works.

\section{Theoretical Background}
\label{theoretical_background}

In this section, we first introduce domains of \gls{qa} systems in Section \ref{tb_domains_of_qa_systems}. Afterwards, question types and architecture of \gls{qa} systems are described in Sections \ref{tb_question_types_of_qa_systems} and \ref{tb_architecture_of_qa_systems} respectively. In Section \ref{tb_knowledge_distillation}, we review the concept of \gls{kd} as a model compression technique. Ultimately, we describe \gls{al} method in Section \ref{tb_active_learning} that aims at reducing the annotation costs.

\vspace*{-0.05in}
\subsection{Domains of \gls{qa} Systems}
\label{tb_domains_of_qa_systems}
\vspace*{-0.05in}

\Gls{qa} systems fall into two general categories of {\em open domain} and {\em closed domain} systems \cite{molla2006special}. Open (or unrestricted) domains aim at answering various types of questions about a diverse set of subjects such as sports, politics, religions, etc. \cite{KOLOMIYETS20115412}. In contrast, closed (or restricted) domains are bound to answer the questions associated with a specific subject. The task of these systems is in general simpler than that of open domain cases because \gls{nlp} models can extract information from a specific domain and utilize its features to predict a suitable answer to a given question \cite{lakshmi2019}. Typically, the model covers the answers to a limited number of questions that are frequently used in a restricted domain \cite{KOLOMIYETS20115412}.

\vspace*{-0.05in}
\subsection{Question Types of \gls{qa} Systems}
\label{tb_question_types_of_qa_systems}
\vspace*{-0.05in}

Different types of classifications of questions are available; however, in a particular semantic category, which has absorbed more attention, questions have been divided into categories of {\em factoid}, {\em list}, {\em definition (or description)}, {\em hypothetical}, {\em causal}, {\em relationship}, {\em procedural}, and {\em confirmation} \cite{KOLOMIYETS20115412}. In English, a factoid question normally starts with a Wh-interrogative word such as ``Who,'' ``What,'' ``When,'' or ``Where'' \cite{KOLOMIYETS20115412}. The answer to such a question is usually an embedded fact within the body of the text that can be either a {\em numerical} or a {\em named entity}. On the other hand, a list question is a type of question with an answer as a list of text entities. Alternatively, an answer to a definition question can be a full sentence about a term used in the body of the question. Furthermore, answering a hypothetical question requires information about a hypothetical event. To answer a causal question, however, the system looks for information or an explanation about an event and the question typically starts with ``Why.'' On the other hand, to answer a relationship question, the system searches for a relationship established between two entities. A procedural question is a type of question with an answer including all the instructions required to fulfill the task mentioned in the question. Lastly, a confirmation question requires a yes or no answer for the event mentioned in the body of the question. Alternatively, questions can be divided into two general categories of factoid and non-factoid. In this case, non-factoid questions, which are more complex to answer, include all question types except for the factoid one mentioned above.

\vspace*{-0.05in}
\subsection{Architectures of \gls{qa} Systems}
\label{tb_architecture_of_qa_systems}
\vspace*{-0.05in}

\begin{figure}[H]
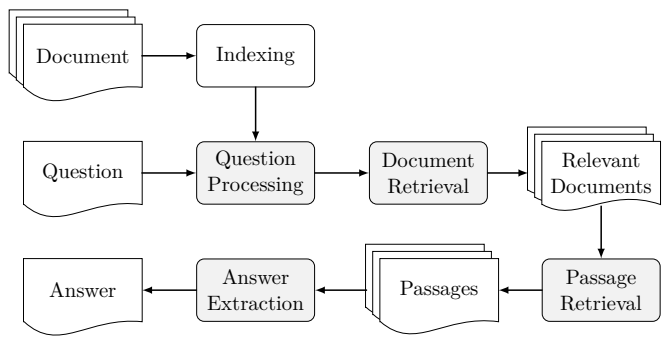

	\begin{center}
		\includestandalone[mode=buildnew,width=0.6\linewidth]{./assets/ir_based_qa_systems_architecture}
	\end{center}
	\caption{The architecture of \gls{ir}-based \gls{qa} systems consists of question processing, document retrieval, passage retrieval, and answer extraction modules.}
	\label{fig:ir_based_qa_systems_architecture}
\end{figure}

\begin{figure}[H]
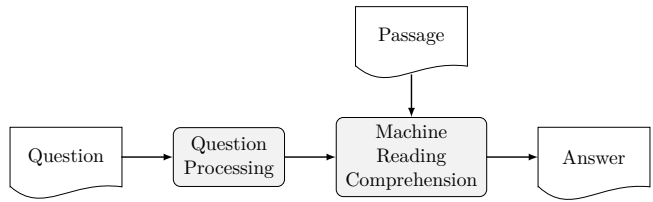

	\begin{center}
		\includestandalone[mode=buildnew,width=0.6\linewidth]{./assets/rc_based_qa_systems_architecture}
	\end{center}
	\caption{The architecture of \gls{rc}-based QA systems consists of question processing and machine reading comprehension modules.}
	\label{fig:rc_based_qa_systems_architecture}
\end{figure}

The architectures of \gls{qa} systems can be divided into \gls{ir}-based or \gls{rc}-based systems \cite{lakshmi2019}. In Figures \ref{fig:ir_based_qa_systems_architecture} and \ref{fig:rc_based_qa_systems_architecture}, a general overview of these architectures is depicted. The \gls{ir}-based systems includes four modules of {\em question processing}, {\em document retrieval}, {\em passage retrieval}, and {\em answer extraction}.

In the question processing module, the required processes are performed on the question body. Semantic and structural relations between the question words are extracted. Then, in the document retrieval module, the documents are ranked. Next, in the passage retrieval module, the most fitting segments of highly ranked documents that are related to the input question are chosen as the candidate passage. Finally, in the answer extraction module, the candidate passages are used to return the most probable answer \cite{KOLOMIYETS20115412}.

In RC-based systems, context paragraphs and their corresponding questions are represented as a series of tokens $P=\{p_1, p_2, p_3, ..., p_n\}$ and $Q=\{q_1, q_2, q_3, ..., q_n\}$ \cite{yu2018qanet}. The goal here is to predict the answer in the form of a span within one of the context paragraphs, $A=\{p_j, ..., p_{j+k}\}$. In such cases, the system is expected to analyze questions and context paragraphs comprehensively to find the best (i.e., the most relevant) answer. Although several different learning methods have been employed in RC-based systems, deep learning methods, in particular, have achieved a higher accuracy \cite{lakshmi2019}.

\vspace*{-0.05in}
\subsection{Knowledge Distillation}
\label{tb_knowledge_distillation}
\vspace*{-0.05in}

An effective technique for model compression, known as knowledge distillation, has recently gained much popularity among researchers. Using \gls{kd}, a compact neural network can be trained in such a way that we achieve the same high accuracy of a much larger network \cite{hinton2015distilling}.

The \gls{kd} architecture is composed of two components, i.e., a {\em student} model and a {\em teacher} model. The teacher component is a large model with high accuracy but with heavy computational costs and a large number of parameters. On the other hand, the student component is a compact model with a smaller number of parameters. The student model mimics the teacher's behavior. However, it is more suitable for deployment due to much lower computational costs. To imitate the behavior of the teacher, the student, along with its own {\em actual labels} ({\em hard target}), also employs the teacher's {\em output logits} ({\em soft target}). As it follows, the loss function consists of adding hard and soft terms:

\begin{equation}
	\label{eq:mrt01}
	L = (1 - \rho) C_{hard}(\vb*{x}, \vb*{y}) + \rho C_{soft}(\vb*{x}, \vb*{q}),
\end{equation}

\begin{equation}
	\label{eq:mrt02}
	C_{hard}(\vb*{x}, \vb*{y}) = - \sum\limits_{i=1}^{K}y_i \log p_i(\vb*{x}),
\end{equation}

\begin{equation}
	\label{eq:mrt03}
	C_{soft}(\vb*{x}, \vb*{q}) = - \sum\limits_{i=1}^{K}q_i \log p_i(\vb*{x}),
\end{equation}

where $C_{hard}$ is the \gls{ce} loss function of the student model and $C_{soft}$ is applied to the softmax of the output of both models. $\rho$ is the weight of the hard and soft cross-entropy losses. $K$ is the number of output classes of $x$. $p_i(x)$ is the softmax output probability of the $i$-th class of the student. The hard target $y$ is a one-hot $K$-dimensional vector. $q$ is a soft target, which is a $K$-dimensional vector. $q_i$ is the {\em tempered softmax probability} for $i$-th class of the teacher model, which is computed as follows \cite{7953145}:

\begin{equation}
	\label{eq:mrt04}
	q_i = \frac{exp(z_i(\vb*{x})/T)}{\sum_{j=1}^{K}{exp(z_j(\vb*{x})/T)}},
\end{equation}

where $z_i(\vb*{x})$ is the pre-softmax output of the teacher model for the $i$-th class. $T$ is the {\em temperature}. When $T$ is large, the class probability distribution will be uniform. In other words, $q$ is a smooth probability distribution containing the correct class information and between-class similarity. Learning these correlations has a massive effect on the performance of the student model. Temperature $T$ controls the importance of the class similarity information during training. When $T$ is greater than 1, small probabilities of non-target classes are emphasized; in that case, the student learns class similarity information more accurately \cite{hinton2015distilling}. 

\vspace*{-0.05in}
\subsection{Active Learning}
\label{tb_active_learning}
\vspace*{-0.05in}

\Gls{al} is a learning method that aims at minimizing the annotation costs without sacrificing the accuracy \cite{Fu2013}. The main purpose of this approach is that if the training algorithm is able to choose more informative data during the learning process, the model can reach almost the same accuracy as a supervised method with a much less amount of data. \Gls{al} approaches are classified into three major categories of {\em membership query synthesis}, {\em stream-based selective sampling}, and {\em pool-based sampling}.

In membership query synthesis, new instances are generated for which an omniscient expert is expected to provide the ground-truth labels. However, those instances may not have a natural distribution, making the annotation difficult even for a human \cite{settles.tr09}. Selective sampling is an alternative approach for synthesizing queries. This approach is also called stream-based (or sequential) \gls{al}. Here, unlabeled instances are firstly sampled by the actual distribution. Then it is decided if the samples should be labeled based on their value \cite{settles.tr09}. The pool-based sampling approach is based on the assumption that we have a small set of labeled and an enormous pool of unlabeled data. The best candidates (i.e., the most informative ones) are selected from the pool by different selection criteria, annotated by an oracle, and added to the labeled dataset. The training process is repeated every time that some labeled samples are added to the training set \cite{settles.tr09,amiriparian2017cast}. All \gls{al} strategies must measure the usefulness of unlabeled data based on some specified criteria, among which the most popular one is the {\em uncertainty measure}. 

\section{Related Works}
\label{related_works}

In this section, we first review the conventional and contemporary machine learning methods for \gls{qa} systems in Sections \ref{rw_deep_learning_based_models} and \ref{rw_semi_supervised_learning}. Then, we compare various \gls{kd} and \gls{al} approaches in Sections \ref{rw_knowledge_distillation} and \ref{rw_active_learning}, respectively.

\vspace*{-0.05in}
\subsection{Deep Learning-Based Models}
\label{rw_deep_learning_based_models}
\vspace*{-0.05in}

In 2016, dynamic chunk reader was presented \cite{yu2016endtoend}. It was able to extract varied length answers; whereas, its predecessor models returned one word or a named entity as the answer for each question. One of the widely used models is called BiDAF \cite{seo2018bidirectional}, which employs \gls{lstm} and bidirectional attention flow networks. To implement the character level embedding, they have applied \glspl{cnn}, and to obtain the vector representation of each word, they have used GloVe \cite{pennington-etal-2014-glove}, a pre-trained word embedding. In 2017, the DrQA model \cite{chen-etal-2017-reading} was introduced. It consists of two modules of {\em document retriever}, which extracts five related documents for each question, and {\em document reader} composed of a bidirectional \gls{lstm} network. {\em Transformer} was introduced in 2017 \cite{NIPS2017_3f5ee243}. Instead of using \glspl{rnn} or \glspl{cnn}, a self-attention mechanism has been used to increase parallelism. Transformers are encoder-decoder-based models that heavily rely on the self-attention mechanism. Despite their overall high accuracy, these models are extremely vulnerable when facing adversarial samples, which results in low accuracy. In 2018, a model with a structure consisting of a {\em sentence selector} connected to a \gls{qa} model was proposed \cite{min-etal-2018-efficient}. The sentence selector computes a selection score for each sentence based on its word-level relevance and semantic similarity to the question. Sentences with the highest scores are selected to be fed to the \gls{qa} model. Additionally, an encoder with a similar structure to DrQA has been used in this model.

QANet is a model which uses \gls{cnn} instead of a recurrent architecture \cite{yu2018qanet}. It was proposed in 2018. The encoder structure in QANet consists of a {\em convolution}, a {\em self-attention}, and a {\em feedforward} layer. After encoding the question and the answer, a standard self-attention mechanism is used to learn the relations between the question and its corresponding answer. The improvement of the learning speed has made QANet a suitable candidate for applying {\em data augmentation}. Accordingly, using neural machine translation (NMT) \cite{luong-etal-2015-effective}, the {\em back-translation} method has also been employed in QANet for the data augmentation purpose. BERT is an extremely popular model, initially released in late 2018 \cite{devlin-etal-2019-bert}. Using bidirectional Transformer encoders, BERT was unsupervised pre-trained on the tasks of masked language modeling (MLM) and next sentence prediction (NSP). It has the capability of being fine-tuned on a wide array of downstream tasks. BERT-like models managed to outperform previous solutions on several \gls{nlp} tasks, especially \gls{qa} tasks. XLNet is another successful architecture, which is based on autoregressive language models. It has been fine-tuned for \gls{qa} by some other models \cite{yang2020xlnet}.

\vspace*{-0.05in}
\subsection{Semi-Supervised Learning}
\label{rw_semi_supervised_learning}
\vspace*{-0.05in}

In the past few years, limited researches have been conducted on semi-supervised \gls{qa} systems. A model called GDAN \cite{yang-etal-2017-semi} was proposed in 2017. This model uses a {\em generator} to make fake questions using a set of unlabeled documents, in addition to the real questions made by a human expert using the same dataset. These generated questions are then fed to a {\em discriminator} that tries to distinguish real questions from fake ones. The learning procedure of both generator and discriminator networks continues until that the discriminator will be unable to recognize the fake questions.  There is another semi-supervised \gls{qa} research introduced in 2018, in which a series of questions corresponding to a specific document structure is generated \cite{dhingra-etal-2018-simple}. The main idea of this research is that the introduction section includes some questions that are elaborately answered in the body of the article. Accordingly, all sentences in the introduction, which is assumed to be the initial 20\% of the document, are regarded as questions $\{q_1, q_2, q_3, ..., q_n\}$, and the remaining 80\% is supposed to include the paragraphs $\{p_1, p_2, p_3, ..., p_m\}$ that contain the answers. Then, the matching $match(p_i, q_i)$ is computed for each given question-paragraph pair. Whenever there is an exact match between the tokens of a question and a paragraph, the matched span is identified as the answer to the question.

In another study, both supervised and unsupervised transfer learning has been used. The focus of the study was on multiple-choice question answering \cite{chung-etal-2018-supervised}. Additionally, in another research conducted in 2018 \cite{min-etal-2017-question}, the transfer learning method was employed for improving the learning ability of the network. In this approach, SQuAD was used as the source dataset for pre-training the model; both WikiQA and SemEval 2016 were used as the target datasets. In \cite{lewis-etal-2019-unsupervised}, unsupervised learning was used for \gls{qa} systems. To generate context, question, and answer triples, some noun phrases and named entity mentions are selected as the candidate answers. Then, these answers are converted to the form of ``fill-in-the-blank" cloze questions and finally translated into natural questions.

In other areas of artificial intelligence, semi-supervised learning is deemed an attractive technique. Many studies have been conducted on semi-supervised learning in word sense disambiguation \cite{Baskaya2016SemisupervisedLW}, temporal relation extraction \cite{Mirroshandel2012TowardsUL}, and image classification \cite{rasmus2015semisupervised,laine2017temporal,tarvainen2018mean,miyato2018virtual}.

\vspace*{-0.05in}
\subsection{Knowledge Distillation}
\label{rw_knowledge_distillation}
\vspace*{-0.05in}

It was shown that \gls{kd} can improve the model generalization and robustness. For instance, using this technique in a \gls{qa} system, the knowledge was transferred from an ensemble teacher to a single student model \cite{hu-etal-2018-attention}. The reinforced mnemonic reader (RMR) is a base model in which attention and reinforcement learning have been integrated \cite{hu2018reinforced}. This model was evaluated on SQuAD, Adversarial SQuAD, and NarrativeQA datasets. In this work, the student was made of a single RMR and the teacher was an ensemble model composed of 12 copies of the base model (i.e., RMR), each having different initial parameters. A two-stage \gls{kd} strategy with multiple teachers was used for web \gls{qa} systems \cite{yang2019model}. These two stages are {\em pre-training} and {\em fine-tuning}. The results of this study showed that this method is performant in generalization. \Gls{skd} was used in \cite{hahn-choi-2019-self}. As it was mentioned before, in \gls{kd}, the knowledge is normally transferred from a large (teacher) model to a small (student) model. However, in \gls{skd}, the source of the knowledge is the student model itself. The results of applying \gls{kd} methods in a study conducted on dialog systems \cite{arora-etal-2019-knowledge} with a dataset named Holl-E demonstrate that imitating the behavior of the teacher model has a significant impact on the student's performance.

Recently, some studies have focused on \gls{kd} using the BERT model as the teacher. The main objective is to create a compact pre-trained (student) model with much fewer parameters and much less inference time than that of the BERT model, but at the same time with competitive accuracy. DistilBERT was presented in 2019 \cite{sanh2020distilbert}. Unlike all previous models, in DistilBERT the \gls{kd} method is performed during the model pre-training stage. In this structure, the BERT\textsubscript{BASE} model is the teacher. The total number of encoder layers of the student is half of the layers of the teacher though their structures are identical. It was demonstrated that using the DistilBERT model, the BERT's size can be reduced by 40\% while preserving 97\% of its language comprehension capabilities \cite{sanh2020distilbert}. TinyBERT is another BERT\textsubscript{BASE} model created by \gls{kd} \cite{jiao-etal-2020-tinybert}. The \gls{kd} method used in this model is called Transformer distillation that is performed in two stages of {\em general distillation} and {\em task-specific distillation}. At the general distillation stage, the BERT model without fine-tuning is used as the teacher, and TinyBERT is taken as the student that imitates the teacher's behavior through the application of Transformer distillation to a general-domain corpus. At the task-specific distillation stage, however, this general TinyBERT model is used for distilling the knowledge. For this purpose, at first, the data augmentation process is performed. Then with the fine-tuned BERT model used as the teacher, \gls{kd} is applied to the resultant dataset. Both stages are necessary for the TinyBERT model to achieve effective performance and generalization. TinyBERT with four layers is 7.5 times smaller than the BERT\textsubscript{BASE} model. Also, in terms of the inference time, it is 9.4 times faster than BERT\textsubscript{BASE}. Nonetheless, it has gained 96.8\% performance of BERT\textsubscript{BASE} applied to the GLUE benchmark. In another study \cite{sun2019patient}, \gls{kd} was used to transfer knowledge from the BERT model as the teacher to a student model. In this work, intermediate layers along with the output of the last layer were used as the medium of transferring knowledge, which showed satisfactory results in several \gls{nlp} downstream tasks. Furthermore, \gls{kd} has also achieved promising results in some other concepts such as multi-task learning \cite{clark-etal-2019-bam,liu2019improving}.

\vspace*{-0.05in}
\subsection{Active Learning}
\label{rw_active_learning}
\vspace*{-0.05in}

\Gls{al} has been widely used in different subtasks of \gls{nlp}. As an example, in a research study focused on \gls{ner}, \gls{al} was applied to a deep learning structure \cite{shen-etal-2017-deep}. The model used two \glspl{cnn} for encoding characters and words, in addition to an \gls{lstm} network as a decoder. The results showed, with the aid of \gls{al} and merely one-fourth of the training dataset, the model achieved 99\% accuracy of the best deep learning models trained on the whole dataset. In \cite{Liu2020ltpan}, using the BERT-CRF model, an uncertainty-based \gls{al} strategy was applied to \gls{ner} and achieved satisfactory results.

Although the combination of \gls{al} and deep learning has been recently applied to other text processing fields such as coreference resolution \cite{li-etal-2020-active-learning}, entity resolution \cite{kasai2019lowresource}, machine translation \cite{liu-etal-2018-learning}, and dependency parsing \cite{mirroshandel-nasr-2011-active}, it has not been used in \gls{qa} tasks. Moreover, \gls{al} has been used in applications such as imbalanced datasets \cite{9093475} and black box attack \cite{li2018queryefficient}.

\section{Proposed Approaches}
\label{proposed_approaches}

We propose an interpolated \gls{kd} method to transfer knowledge to the model and reduce its complexity, and \gls{al} strategies to minimize the labeled data requirement. We combine these two approaches to building a small model that gains the high accuracy of a complex model trained on a large corpus, using only a small training dataset. Our approaches are explained in detail in Sections \ref{approach_1} and \ref{approach_2}.

\vspace*{-0.05in}
\subsection{Knowledge Distillation for \gls{qa}}
\label{approach_1}
\vspace*{-0.05in}

Pre-trained models such as BERT have achieved outstanding results in several \gls{nlp} tasks. However, as it was mentioned before, available \glspl{dnn} are extremely complex computational-wise, which makes them unfit for practical applications. Our proposed approach to tackle this issue in \gls{qa} systems is to apply \gls{kd} methods to such networks.

\begin{figure}[H]
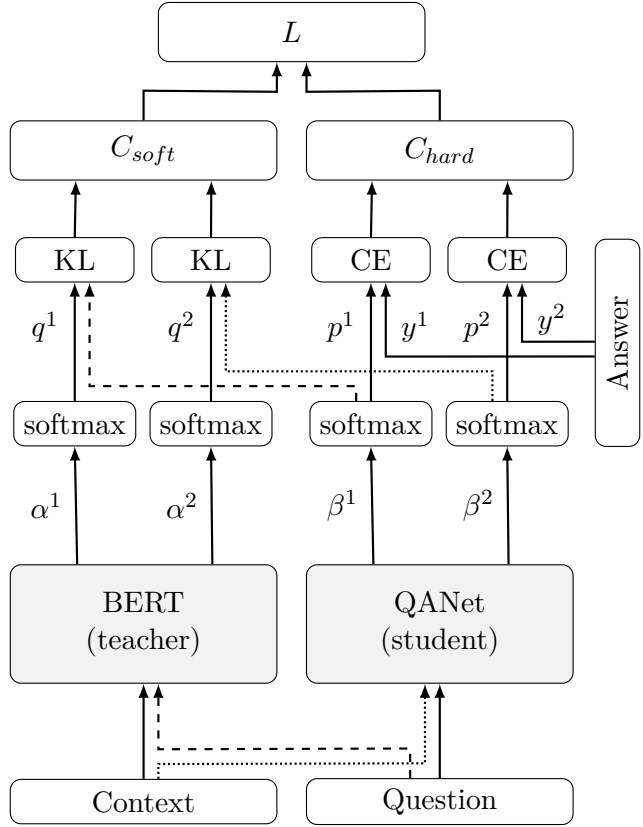

	\begin{center}
		\includestandalone[mode=buildnew, width=0.55\linewidth]{./assets/approach_1_knowledge_distillation}
		\caption{Our proposed model structure employs an interpolated \gls{kd} with BERT\textsubscript{LARGE} as the teacher model and QANet as the student model. $C_{soft}$ is the KL divergence error function that transfer knowledge from BERT\textsubscript{LARGE} to QANet. $C_{hard}$ is the CE loss function of the QANet model. $L$ is the weighted combination of $C_{soft}$ and $C_{hard}$.}
		\label{fig:approach_1_knowledge_distillation}
	\end{center}
\end{figure}

The proposed model structure is depicted in Figure \ref{fig:approach_1_knowledge_distillation}. In this structure, BERT\textsubscript{LARGE} \cite{devlin-etal-2019-bert} and QANet \cite{yu2018qanet} are used as the teacher and the student model, respectively. Instead of using \gls{rnn}, QANet has been designed to use \gls{cnn} in addition to several self-attention layers. As a result of this change, QANet has been shown to have a lower training and inference time in comparison with earlier \gls{qa} models. The formulations used in Figure \ref{fig:approach_1_knowledge_distillation} are as follows:

\begin{equation}
	\begin{aligned}
		L & = (1-\rho) C_{hard} + \rho C_{soft} 
	\end{aligned}
	\label{eq:kd_loss_l}
\end{equation}

\begin{equation}
	\begin{aligned}
		C_{hard} & = \sum_{i=1, 2} \text{CE}(\text{softmax}(\vb*{\beta^i}), \vb*{y^i}) 
	\end{aligned}
	\label{eq:kd_loss_chard}
\end{equation}

\begin{equation}
	\begin{aligned}
		C_{soft} & = T^2 \sum_{i=1,2} \text{KL}(\vb*{p^i}, \vb*{q^i}) 
	\end{aligned}
	\label{eq:kd_loss_csoft}
\end{equation}

\begin{equation}
	\begin{aligned}
		\vb*{q^i} & = \text{softmax}(\vb*{\alpha^i}/T) 
	\end{aligned}
	\label{eq:kd_loss_qi}
\end{equation}

\begin{equation}
	\begin{aligned}
		\vb*{p^i} & = \text{softmax}(\vb*{\beta^i}/T) 
	\end{aligned}
	\label{eq:kd_loss_pi}
\end{equation}

The employment of \gls{kd} in this work has been done as follows:

In standard \gls{qa} models, the cross-entropy loss function is based on Equation (\ref{eq:mrt05}). This term is shown as $C_{hard}$ in Figure \ref{fig:approach_1_knowledge_distillation}.

\begin{equation}
	\label{eq:mrt05}
	\begin{aligned}
		L_{CE} = -\sum_{k=1}^{m} \sum_{l=1}^{m} \vb*{y_k^1} \log p^1(k) + \vb*{y_{l}^2} \log p^2(l \lvert k) 
	\end{aligned}
\end{equation}

$y^1$ and $y^2$ are one-hots for the start and end answer tokens. $m$ is the length of the paragraph. To apply \gls{kd}, \gls{kl} divergence error function is added to the cross-entropy error function, according to Equation (\ref{eq:mrt06}). This term is shown as $C_{soft}$ in Figure \ref{fig:approach_1_knowledge_distillation}.

\begin{equation}
	\label{eq:mrt06}
	\begin{aligned}
		L_{KD} = \text{KL}(p \Vert q)                                 
		= - \sum_{k=1}^{m} \sum_{l=1}^{m} p^1(k) \log [p^1(k)/q^1(k)] 
		+ p^2(l \vert k) \log [p^2(l \vert k) / q^2(l \vert k)]       
	\end{aligned}
\end{equation}

$q$ is the probability distribution of the start and end of the answer, which is extracted from the teacher model. Additionally, log-of-softmax is used to compute $p$ and $q$. Below, we briefly describe the architecture of both teacher and student models used in this study. 

\subsubsection{The BERT Model}

The \gls{bert} \cite{devlin-etal-2019-bert} is a language model capable of being used as a pre-trained model. The BERT's architecture is based on the encoder structure of Transformers. Instead of \gls{cnn} and \gls{rnn} components, the Transformer architecture comprises a number of attention and self-attention layers, with the aim of increasing parallelism \cite{NIPS2017_3f5ee243}. BERT is trained on a masked language modeling task, which allows bidirectional training (i.e., simultaneous consideration of both left and right contexts) in the model. It has been shown that in many \gls{nlp} downstream tasks, we can achieve much improved results by just adding a few layers to the pre-trained BERT model.

\subsubsection{The QANet Model}

\begin{figure}[H]
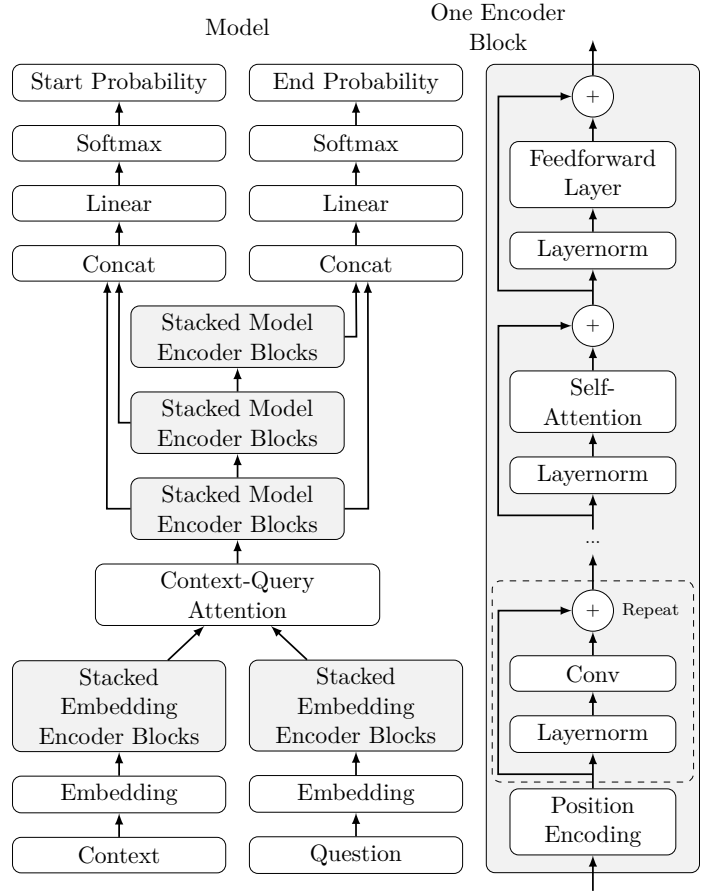

	\centering
	\includestandalone[mode=buildnew, width=0.6\linewidth]{./assets/qanet_model_architecture}
	\caption{The QANet model architecture (left) adopted from \cite{yu2018qanet} consists of multiple encoder blocks. A single encoder block (right) includes a convolution, a self-attention, and a feed-forward layer.
	}
	\label{fig:qanet_model}
\end{figure}

The architecture of QANet \cite{yu2018qanet}, which is shown in Figure \ref{fig:qanet_model}, includes five main layers: {\em embedding}, {\em embedding encoder}, {\em context-query attention}, {\em model encoder}, and {\em output} layer. The convolutional and self-attention networks in the embedding encoder and model encoder layers process the input tokens in parallel, which leads to a significant increase in the performance of QANet in contrast with other models.

Self-attention in this model is similar to that of Transformer. The embedding layer takes advantage of GLoVe \cite{pennington-etal-2014-glove} for word embedding and \glspl{cnn} for character embedding, both of which are connected by their outputs. The embedding encoder layer consists of a stack of encoders. An encoder block with its internal components is shown on the right side of Figure \ref{fig:qanet_model}. These components include a convolutional, a self-attention, and a feed-forward layer. The size of the kernel is set to 7. The number of filters and convolution layers are set to 128 and 4, respectively. Similar to Transformers, the self-attention structure uses a multi-head attention module. All these layers (i.e., convolution, self-attention, and feed-forward) are inside a separate residual block. The structure of each model encoder layer is similar to the block on the right-hand side of Figure \ref{fig:qanet_model}. However, there are two convolution layers in each block, and there are a total of seven encoder blocks in each layer.

There are three stacks of model encoders in QANet, and the parameters are shared between all these encoders. In the output layer, the result of the three stacks of model encoders is used to compute the start and end probabilities of the answer span. For each token of the paragraph, the probability values of the start and end of the span are computed using a linear network and softmax.

\subsubsection{Tokenizers Alignment}

An issue with the \gls{kd} method in the proposed architecture is that the tokenization algorithms used by the student and teacher models are not the same. Spacy and WordPiece are the two different tokenizers used by QANet and BERT, respectively. As a result of this issue, the size of the output of these models is different. It should be noted that some words are divided into multiple tokens using WordPiece. In such cases, a (\texttt{\#\#}) sign will be attached to all the sub-words except for the first one, indicating that those sub-words should be appended to their preceding token. To apply the \gls{kd} loss function, the output of the models must have the same dimension. To tackle this issue, we propose the following two approaches:

{\bf Rule-Based Approach. }The token alignment algorithm that we have used consists of two main steps of (1) finding exactly matched tokens and (2) finding partially matched tokens. Before performing any comparison, all tokens are converted into lower-case characters of ASCII encoding. For example, the word \texttt{Accommodation} is tokenized as \texttt{[acc, \#\#ommo, \#\#dation]} by the BERT tokenizer. After undergoing the mentioned conversion, these tokens are updated to \texttt{[acc, \#\#ommo, \#\#dation]}. In such a case, the same conversion is carried out by the QANet tokenizer but results in \texttt{[accommodation]}. Then, the two mentioned steps are performed as follows:

(1) If a QANet token is exactly matched by one of the BERT's tokens, the start-end logits computed by the BERT model are retained.

(2) When there are no exact matched tokens, the algorithm skips the first step in favor of the second one. In this case, one of the following two conditions holds: Either, (a) the initial character set of the BERT token is a subset of characters in the QANet token, as shown in Figure \ref{tokenizer_bert_subset_of_qanet}; or (b) this subset relation is reversed, as shown in Figure \ref{tokenizer_qanet_subset_of_bert}.

(2-a) In the majority of cases, when two tokens are not the same, the smaller one is the token extracted by the BERT tokenizer. For instance, the string \texttt{Nuclear Astrophysics} is tokenized and processed as follows:

\begin{figure}[H]
\begin{center}
\begin{BVerbatim}
QANet Tokens    BERT Tokens
1 nuclear         nuclear
2 astrophysics    astro
3 .               ##physics
4                 .
\end{BVerbatim}
\end{center}
\caption{An example where there are no exact matched tokens, and the BERT token (i.e., \texttt{astro}) is a subset of the QANet token (i.e., \texttt{astrophysics}).}
\label{tokenizer_bert_subset_of_qanet}
\end{figure}

The BERT token \texttt{astro} is matched by the first part of the QANet token \texttt{astrophysics}. The algorithm continues over the list of the BERT tokens and finds the token \texttt{physics}, which is a sub-word and should be concatenated with its previous token, \texttt{astro}. This procedure continues until the concatenated string of tokens matches the QANet token, i.e., \texttt{astrophysics}. In this example, with one forward step in the BERT token list, the concatenated string matches the QANet token. In this case, the start and end logits of the token \texttt{astro} are retained to be used later for computing the loss function of \gls{kd}; and that of the token \texttt{physics} is ignored. The next comparison is between the two (\texttt{.}) characters, which results in an exact match and retaining the BERT's start and end logits for this token. This operation continues until the last member of the BERT token list is processed.

(2-b) If the QANet token is the smaller one (for example in the tokenization of the string \texttt{cannot understand}) the procedure is performed as follows:

\begin{figure}[H]
\centering
\begin{BVerbatim}
QANet Tokens    BERT Tokens
1 can             cannot
2 not             understand
3 understand
\end{BVerbatim}
\caption{An example where there are no exact matched tokens, and the QANet token (i.e., \texttt{can}) is a subset of the BERT token (i.e., \texttt{cannot}).}
\label{tokenizer_qanet_subset_of_bert}
\end{figure}

Comparing the QANet token \texttt{can} and token \texttt{cannot} of BERT determines how much we should move forward in the token list of QANet. In this example, the algorithm proceeds to the next token, and a new string is created by concatenation of that token and its preceding token, i.e., \texttt{can}. The resultant string is then compared against the current BERT token. In this case, moving just one token ahead results in an exact match. Similar to (2-a), here the start and end logits of the token \texttt{can} are retained to be used later for calculating the loss function of \gls{kd}; and that of the token \texttt{not} is ignored. In this example, the algorithm will next find another exact match for token \texttt{understand}.

{\bf Interpolation Approach. }As it was explained in step (2) of the rule-based approach, in cases where we do not initially have an exact match, we retain the start and end logits of the first sub-token and ignore those of other sub-tokens. This results in losing some useful information. To overcome this issue, we have designed another approach based on the {\em interpolation} of the student logits. The goal is to make the dimension of the student context logits to become equal to that of the teacher. This approach retains the information content of all sub-token logits and results in an extensive knowledge transfer between the teacher and student models. In this work, we perform both linear and cubic interpolation methods on the original context vector of the student to generate new logits required for resizing the student's vector. For transferring the knowledge, a \gls{mse} loss function is added to the main loss function to minimize the distance between the interpolated student vector and that of the teacher. The new loss function is as follows:

\begin{equation}
	\label{eq:loss_function_after_interpolation}
	\begin{multlined}
		L = (1-\rho) C_{hard} + \rho C_{soft}
		+ MSE(stdt_{intrpl}, tchr)
	\end{multlined}
\end{equation}

\vspace*{-0.05in}
\subsection{Active Learning for \gls{qa}}
\label{approach_2}
\vspace*{-0.05in}

\Gls{al} is an efficient way to reduce the required time for creating a training dataset. In this research, we have used the pool-based \gls{al} method. The experiments have been performed on the SQuAD v1.1 dataset.

As it is shown in Algorithm \ref{algorithm:approach_2_active_learning}, at first, all samples of the training dataset are considered to be unannotated. Then, one percent of the dataset is selected to be used for training the model. In this experiment, the chosen model is BERT\textsubscript{BASE} which is trained for two epochs. Then, 10\% of the rest of the unlabeled dataset is selected to be added to the current training dataset using the following strategies. This procedure continues until all unlabeled samples are exhausted.

\vspace*{-0.05in}
\begin{center}
	\begin{algorithm}[H]
		\caption{Pool-based \gls{al} approach}
		\label{algorithm:approach_2_active_learning}
		\SetAlgoLined
		\KwIn{
			Unlabeled data pool $\mathcal{U}$, labeled data set $\mathcal{L}$, most informative unlabeled samples $\vb{x^{*}}$, \gls{al} sampling $\phi(\cdot, \cdot)$
		}
		$\vb{x^{*}} \longleftarrow \text{arg} \max_{\vb*{x} \in \mathcal{U}} \phi(\vb{x}, 1\%)$\;
		$\mathcal{L} \longleftarrow label(\vb{x^{*}})$\;
		$\mathcal{U} \longleftarrow \mathcal{U} \setminus \vb{x^{*}}$\;
		\Repeat{$\lvert\mathcal{U}\rvert = 0$}{
			$train\_model(\mathcal{L})$\;
			$\vb{x^{*}} \longleftarrow \text{arg} \max_{\vb*{x} \in \mathcal{U}} \phi(\vb{x}, 10\%)$\;
			$\mathcal{L} \longleftarrow \mathcal{L} \cup label(\vb{x^{*}})$\;
			$\mathcal{U} \longleftarrow \mathcal{U} \setminus \vb{x^{*}}$\;
		}
	\end{algorithm}
\end{center}
\vspace*{-0.05in}

Most data sampling strategies are based on some uncertainty criteria. Next, we describe the strategies that we have used in this work.

\subsubsection{Least Confidence}

The most widely used strategy of \gls{al} is the {\em least confidence sampling} \cite{settles.tr09}. The algorithm selects those instances that have the least confidence (i.e., based on our model) for labeling. This method can be simply employed in probabilistic models. For example, in a probabilistic binary classification model, instances with a probability value around 0.5 are the ones in which the model has the least confidence.

The output of the \gls{qa} systems that we are interested in is a span extending from the start to the end of the answer tokens.  For each question, the model returns multiple answer spans, among which the span with the highest probability value will be selected. In each cycle, a fixed number (e.g., 10\%) of questions whose selected answer has the least probability value are selected. The calculations are performed using Equations (\ref{eq:mrt08}) and (\ref{eq:mrt08_a}).

\begin{equation}
	\label{eq:mrt08}
	\begin{aligned}
		\vb*{x^{*}} = \text{arg} \max_{\vb*{x}} \big[ 1 - p(A^{\string^} \vert \vb*{x}) \big] 
	\end{aligned}
\end{equation}

\begin{equation}
	\label{eq:mrt08_a}
	\begin{aligned}
		A^{\string^} & = \text{arg} \max_{A} p(A \vert \vb*{x}) 
	\end{aligned}
\end{equation}

$A$ is the answer set returned by the model for a question. For each instance, $\vb*{x}$, $A^{\string^}$ is the answer with the highest probability value given by the model. In this approach, the selected answer with the least probability value is chosen as the least confident instance, denoted by $\vb*{x^*}$. This instance is presumed to contain the highest information content of all.

\subsubsection{Margin}

Another option that can be used for data sampling is the {\em margin} criterion. In this method, the probability difference between the two most probable labels is calculated. This difference shows that samples with a larger margin are easier to be classified by the model. That is because the classifier is more confident about those labels. Conversely, the classifier is less confident about those labels that have a smaller margin; therefore, knowing the actual label of such instances helps the model discriminate them more effectively. For applying this criterion to \gls{qa} systems, the difference between the two most probable answers returned for each question is taken as the margin. This margin is calculated by Equation (\ref{eq:mrt09}), in which $A_1^{\string^}$ and $A_2^{\string^}$ respectively denote the first two most probable answer to question $\vb*{x}$. Here, in each \gls{al} cycle, a subset of questions with the highest margin, denoted by $\vb*{x^{*}}$, are selected to be added to the training dataset.

\begin{equation}
	\label{eq:mrt09}
	\vb*{x^{*}} = \text{arg} \max_{\vb*{x}} \big[ p(A^{\string^}_1 \vert \vb*{x}) - p(A^{\string^}_2 \vert \vb*{x}) \big]
\end{equation}

\subsubsection{Entropy}

When there exist a large number of labels, the margin sampling method practically ignores many labels. In such cases, it only considers the first two labels. In this situation, the sampling method based on {\em entropy}, which is calculated by Equation (\ref{eq:entropy}), is more suitable for detecting uncertainty. $A_i^{\string^}$ denotes the $i$-th most probable answer returned for question $\vb*{x}$.

\begin{equation}
	\label{eq:entropy}
	\vb*{x^{*}} = \text{arg} \max_{\vb*{x}} \big[ - \sum_{i} p(A^{\string^}_i \vert \vb*{x}) \log{(A^{\string^}_i \vert \vb*{x})} \big]
\end{equation}

For applying this method to \gls{qa} systems, the first five most probable answers for each question are selected as the candidate answers by the BERT model. After calculating the entropy for these candidates, the samples with the highest entropy are selected to be added to the training dataset.

\subsubsection{Clustering Method}

{\em Clustering} is another approach used in our study for data sampling. For this purpose, first, some samples are selected from the unlabeled dataset pool by the least confidence approach. If $k$ instances are to be selected for labeling, we initially choose $3 \times k$ instances based on the least confidence criterion as our candidates. Then, for clustering, questions are encoded with the \gls{use} \cite{cer-etal-2018-universal}, and using the $k$-means algorithm and based on the Euclidean distance measure, those candidates will be grouped into 10 clusters. To select final $k$ samples, each cluster is sampled proportional to the number of its members. Selected instances are annotated and added to the current labeled dataset. Then the model is re-trained on the resulting dataset. This procedure continues until our unlabelled data are exhausted.

\section{Experimental Results}
\label{experimental_results}

In this section, to assess the performance of our proposed approaches, we explain the experiments\footnote{Our source code is publicly available at: https://github.com/mirbostani/QA-KD-AL} we have conducted and analyze their results in detail.

\vspace*{-0.05in}
\subsection{Datasets}
\vspace*{-0.05in}

Over the past decades, many datasets have been proposed for \gls{qa} tasks. In this research, for the evaluation purpose, we have used two datasets, SQuAD v1.1 and Adversarial SQuAD, which are depicted in Table \ref{table:dataset} and discussed next.

\vspace*{-0.05in}
\begin{center}
	\begin{table}[H]
		\centering
		\begin{tabular*}{\linewidth}{@{} l @{\extracolsep{\fill}} rrr @{}}
			\hline
			\multicolumn{1}{@{} l}{\bfseries Dataset}
			&
			\multicolumn{1}{r}{\bfseries Documents}
			&
			\multicolumn{1}{r}{\bfseries Paragraphs}
			&
			\multicolumn{1}{r @{}}{\bfseries Questions} \\
			\hline
			SQuAD v1.1: Train & 442 & 18,896 & 87,599 \\
			SQuAD v1.1: Development & 48 & 2,067 & 10,570 \\
			Adversarial SQuAD: ADDSENT & 48 & 3,358 & 3,560 \\
			Adversarial SQuAD: ADDONESENT & 48 & 1,585 & 1,787 \\
			\hline
		\end{tabular*}
		\caption{Statistics of the \gls{squad} v1.1 and Adversarial SQuAD datasets.}
		\label{table:dataset}
	\end{table}
\end{center}
\vspace*{-0.05in}

{\bf SQuAD. }The \gls{squad} v1.1, released in 2016 \cite{rajpurkar-etal-2016-squad}, contains 107,785 question-answer pairs on 536 articles extracted from Wikipedia. In SQuAD v1.1, the answer to each question is a span of the text from the corresponding reading passage. This dataset has provided the ground for significant progress in building more accurate \gls{qa} systems in recent years.

{\bf Adversarial SQuAD. }In 2017, an adversarial dataset was built on top of SQuAD v1.1 \cite{jia2017adversarial}. Its training set has remained unchanged; however, some paragraphs of its validation set have been extended by some adversarial sentences. This dataset provides three different methods for generating adversarial sentences. The first method is called ADDSENT, in which a few adversarial sentences analogous to the given question are generated and appended to the paragraph that contains the answer to that question. In the second method, called ADDONESENT, a fixed sentence is added to all paragraphs. In the last approach, named ADDANY, adversarial sentences are generated and appended to all paragraphs, regardless of grammatical considerations. Therefore, using this dataset, the robustness of \gls{qa} models can be evaluated.

\vspace*{-0.05in}
\subsection{Evaluation Metrics}
\vspace*{-0.05in}

Two common measures used for the evaluation of \gls{qa} systems are the {\em F1-score} and {\em exact match} (EM). To evaluate the performance of the system, its predictions and the ground truth answers are treated as a bag of tokens. The F1-score measures the average overlap between the predicted and the ground truth answers. It is computed for each given question and in the end, is averaged over all the questions. EM measures the percentage of those predictions that exactly match the ground truth answers. Here, when a prediction matches its ground truth answer, EM is set to 1; otherwise, it is set to 0 \cite{rajpurkar-etal-2016-squad}.

\vspace*{-0.05in}
\subsection{Applying Knowledge Distillation}
\vspace*{-0.05in}

To apply \gls{kd}, we used QANet as our student model because it does not use \glspl{rnn} in its architecture; therefore, it has a much lower training and inference time as opposed to other earlier proposed models. Furthermore, BERT\textsubscript{LARGE} (uncased version) was used as our teacher model. It has been suggested that for training the BERT model, the hyperparameters of the model can be set to one of the following learning rates: \num{2e-4}, \num{3e-4}, and \num{5e-4}. In our experiment, we set the rate to \num{5e-4}. The maximum tokens length, which is the maximum length of the input to the model after tokenization, was set to 384. We also utilized the Pytorch framework for implementation. $\rho$ parameter, the coefficient of the soft loss function, was set to 0.7, and the temperature $T$ was set to 10. The model was trained for 30 epochs in a batch size of 14 samples.

\begin{figure}[H]
	\centering
	\begin{subfigure}{.5\textwidth}
		\centering
		\includegraphics[width=.95\linewidth]{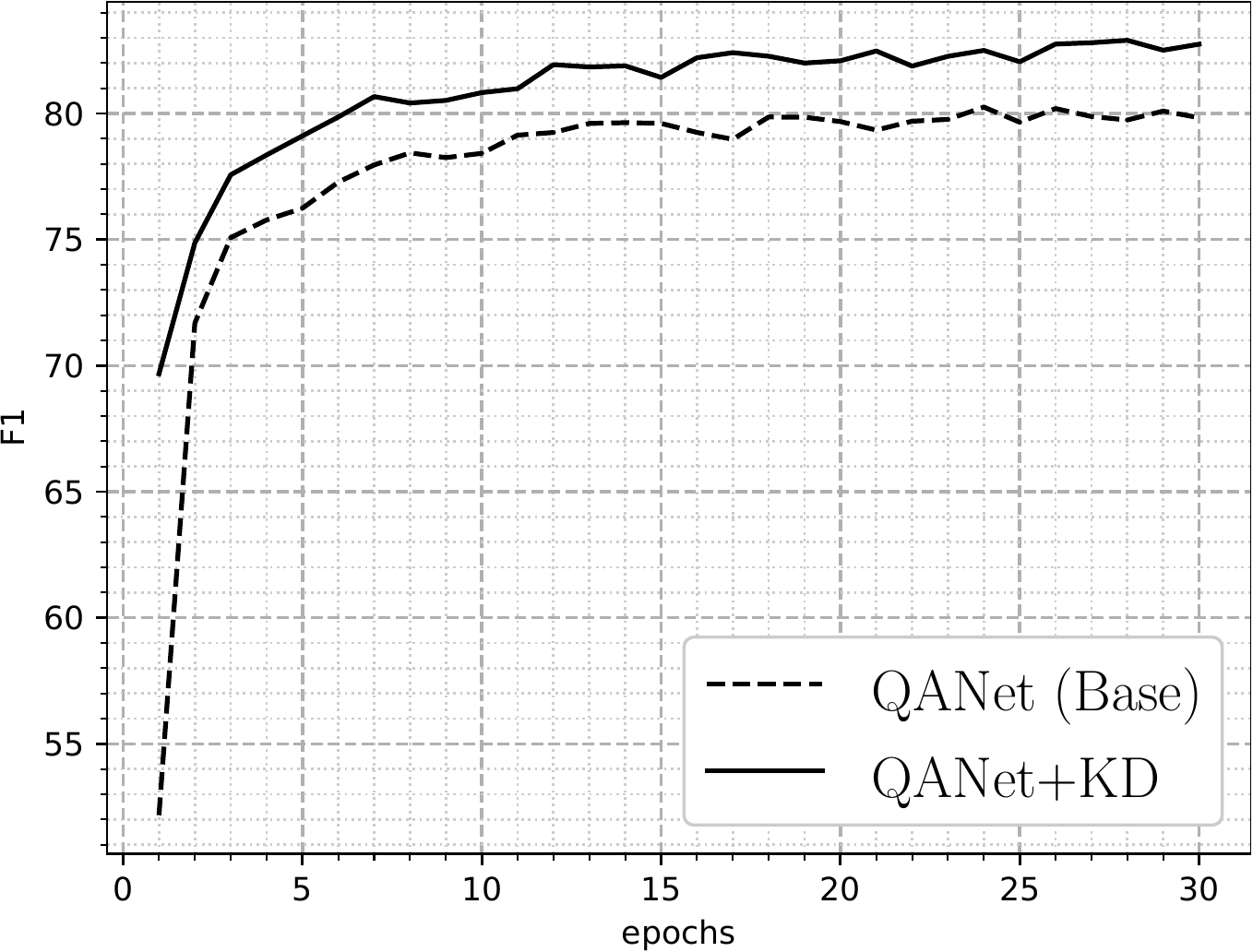}
		\caption{F1-score of QANet versus QANet+KD.}
		\label{fig:qanet_kd_f1}
	\end{subfigure}%
	\begin{subfigure}{.5\textwidth}
		\centering
		\includegraphics[width=.95\linewidth]{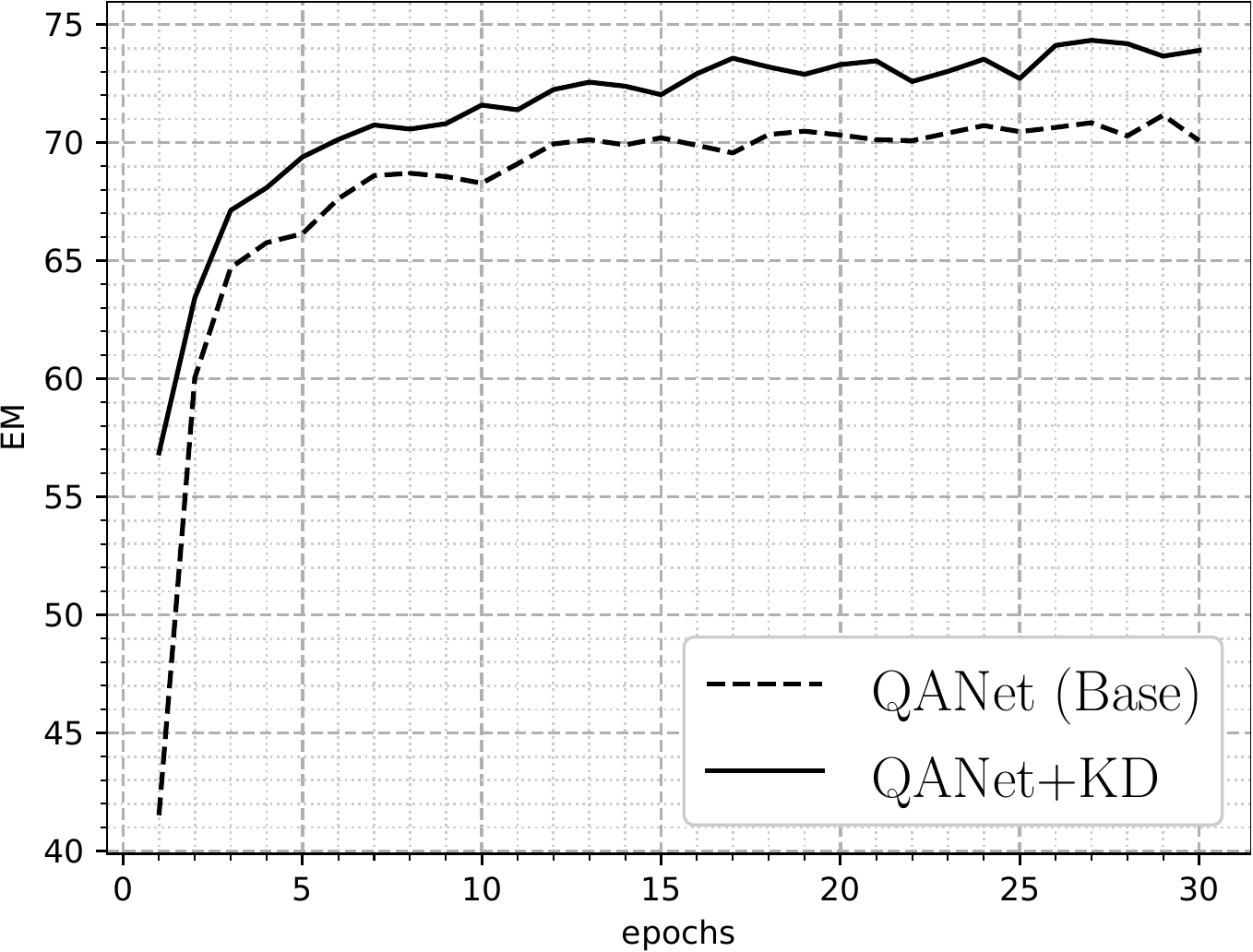}
		\caption{EM of QANet versus QANet+KD.}
		\label{fig:qanet_kd_em}
	\end{subfigure}
	\caption{Performance comparison between QANet before and after applying \gls{kd}.}
	\label{fig:qanet_kd_f1_em}
\end{figure}

As it is demonstrated in Figure \ref{fig:qanet_kd_f1_em}, distilling the knowledge from BERT to QANet increases by around 3.00 F1-score and EM. 

Table \ref{table:performance_on_squad} shows the performance of various combinations of our proposed model in comparison with other related models, using F1-score and EM measure. QANet is the base model used in our study, and QANet+KD is the model on which \gls{kd} has been applied by adding the KL loss function to the model and using the rule-based alignment technique. The QANet+KD+Linear model has a similar model implementation as the QANet+KD with the addition of linear interpolation. Furthermore, the QANet+KD+Cubic model is similar to QANet+KD+Linear except that it utilizes cubic interpolation. The results of our experiments on the SQuAD v1.1 dataset show an improvement of 3.50 and 4.00 percentage points in F1-score and EM of the model, respectively, resulted from \gls{kd} over the base model.

% \vspace*{-0.05in}
\begin{center}
	\begin{table}[H]
		\centering
		\begin{tabular*}{\linewidth}{@{} l @{\extracolsep{\fill}} rr @{}}
			\hline
			\multicolumn{1}{@{} l}{\bfseries Model} &
			\multicolumn{1}{r}{\bfseries F1} &
			\multicolumn{1}{r @{}}{\bfseries EM} \\
			\hline
			BERT\textsubscript{LARGE} (Teacher) \cite{devlin-etal-2019-bert} & 93.15 & 86.91 \\
			\hline
			BERT\textsubscript{BASE} \cite{devlin-etal-2019-bert} & 88.34 & 81.00 \\
			DistilBERT\textsubscript{6} \cite{sanh2020distilbert} & 86.90 & 79.10 \\
			DistilBERT\textsubscript{4} \cite{sanh2020distilbert} & 81.20 & 71.80 \\
			TinyBERT\textsubscript{6} \cite{jiao-etal-2020-tinybert} & 87.50 & 79.70 \\
			TinyBERT\textsubscript{4} \cite{jiao-etal-2020-tinybert} & 82.10 & 72.70 \\
			\hline 
			QANet (Base) \cite{yu2018qanet} & 80.09 & 71.16 \\
			QANet+KD (Proposed) & 83.01 & 73.94 \\
			QANet+KD+Linear (Proposed) & 83.02 & 74.14 \\
			QANet+KD+Cubic (Proposed) & \bfseries 83.51 & \bfseries 75.20 \\
			\hline
		\end{tabular*}
		\caption{Performance of various models on SQuAD v1.1 dataset. BERT\textsubscript{LARGE} and QANet are the teacher and the student model, respectively. QANet+KD is our proposed model after applying \gls{kd}, and QANet+KD+Linear/Cubic are the models with interpolated \gls{kd}. The best performance results, specified in bold, against QANet (Base) belong to our QANet+KD+Cubic model, which outperforms both DistilBERT\textsubscript{4} and TinyBERT\textsubscript{4} and is comparable with these models with six layers.}
		\label{table:performance_on_squad}
	\end{table}
\end{center}
% \vspace*{-0.05in}

One of the problems with large pre-trained language models is their intrinsic computational complexity. To further investigate this issue, we compared the number of parameters and the inference time of our models with other related models. As it is shown in Table \ref{table:param_speedup}, our approach does not change the parameters and inference time of the base model; however, it is capable of improving the performance of the model. Accordingly, choosing the base model directly affects the total number of parameters and inference time of the proposed model.

% \vspace*{-0.05in}
\begin{center}
	\begin{table}[H]
		\centering
		\begin{tabular*}{\linewidth}{@{} l @{\extracolsep{\fill}} rr @{}}
			\hline
			\multirow{2}{*}{{\bfseries Model}} 
			& \multicolumn{1}{r}{{\bfseries \#Params}} 
			& \multirow{1}{*}{{\bfseries Speedup}} \\
			& \multicolumn{1}{r}{{\bfseries (Millions)}} 
			& \multicolumn{1}{r @{}}{{\bfseries (batches/second)}} \\
			\hline
			BERT\textsubscript{BASE} \cite{devlin-etal-2019-bert} & 110.0 & $1.0\times$ \\
			\hline
			DistilBERT\textsubscript{6} \cite{sanh2020distilbert} & 66.0 & $2.0\times$ \\
			DistilBERT\textsubscript{4} \cite{sanh2020distilbert} & 52.2 & $3.0\times$ \\
			TinyBERT\textsubscript{6} \cite{jiao-etal-2020-tinybert} & 67.0 & $2.0\times$ \\
			TinyBERT\textsubscript{4} \cite{jiao-etal-2020-tinybert} & 14.5 & $9.4\times$ \\
			\hline 
			QANet (Base) \cite{yu2018qanet} & 1.3 & $2.0\times$ \\
			QANet+KD (Proposed) & 1.3 & $2.0\times$ \\
			\hline
		\end{tabular*}
		\caption{Number of parameters and speed comparison between our proposed model and other distilled models on SQuAD v1.1 dataset.}
		\label{table:param_speedup}
	\end{table}
\end{center}
% \vspace*{-0.05in}

Note that DistilBERT and TinyBERT are pre-trained models that cannot be trained on a system with limited resources; however, due to the small number of parameters of our proposed model, it can be fully trained on such systems. Although the total number of parameters of the resulted model is about 9\% of the 4-Layer TinyBERT parameters, its F1-score and EM are about 1.40 and 2.50 higher, respectively. Additionally, this model has outperformed the 4-Layer DistilBERT by 2.30 and 3.40 in F1 and EM, respectively, while using 2.5\% of the total number of parameters in DistilBERT. Our model has also achieved around 95\% performance of the 6-Layer TinyBERT and DistilBERT models, using only 2\% of their total number of parameters.

We have validated our results using the {\em bootstrap resampling} technique, a statistical hypothesis testing method, to determine whether there is a significant difference between the means of the two models' predictions. Firstly, as our sample set, 10\% of the evaluation dataset, represented as $X$, was randomly selected and fed to both models. Considering EM as our evaluation metric, the difference between the performance of the initial model before and after applying \gls{kd} on $X$ was calculated as $\delta(X)$. To determine whether the null hypothesis, $H_0: \delta(X) \leq 0$, should be rejected, we must check whether or not $\text{\textit{p}-value} < \alpha$, where $\alpha$ is the {\em significance level}, and $\text{\textit{p}-value}$ is a {\em conditional probability}, based on the null hypothesis $H_0$. For calculating $\text{\textit{p}-value}$, $\delta(X)$ should be resampled with replacement $B$ times to create numerous $\text{\textit{k}-sized}$ sets, where $k$ is the size of $\delta(X)$. Assigning $\alpha$ to $0.05$ and $B$ to $100000$, our calculated $\text{\textit{p}-value}$ is $0.035$ which rejects the null hypothesis and shows the models' performance is statistically significant.

\vspace*{-0.05in}
\subsection{Applying Active Learning}
\vspace*{-0.05in}

We have also applied \gls{al} to the BERT\textsubscript{BASE} model to evaluate the impact of this technique on the volume of required labeled data and the performance of this model. The chosen values for hyperparameters of the model are as follows. The base model of our study was BERT\textsubscript{BASE} (uncased version), the learning rate was set to $\num{5e-4}$, and the maximum token length was set to 384. The BERT\textsubscript{BASE} model was initially fine-tuned for only two epochs. That is because increasing the number of epochs reduces the accuracy of the model on the validation dataset \cite{devlin-etal-2019-bert}. In this experiment, the Pytorch framework was used for implementation. Initially, 1\% of the training dataset was randomly chosen for fine-tuning the BERT\textsubscript{BASE} model; the remaining 99\% of the training data was assumed to be unlabeled. Then, in each step, according to the sampling strategies proposed in Section \ref{approach_2}, in each cycle, 10\% of the remaining samples was added to the current labeled samples used for training. In each cycle, the model was again fine-tuned on the newly compiled dataset. This process was repeated until the model was fully trained on the whole dataset.

In Table \ref{table:active_learning_em}, the impact of various selection strategies on the EM measure is demonstrated. RAND denotes the random sampling strategy, LC stands for the least confidence, EN is entropy, M denotes the margin sampling, and LC-CL is our proposed clustering method. The results of our experiments indicate that the performance of all the sampling methods that we have used outperform the random sampling strategy. Moreover, among these sampling methods, the least confidence strategy has achieved the best results. Using the least confidence strategy and only 20\% the training dataset, the model can achieve 93.83\% EM of the case in which we employ the supervised method and the whole dataset. Additionally, the model can achieve 98.08\% EM with only 40\% of the training dataset.

% \vspace*{-0.05in}
\begin{center}
	\begin{table}[H]
		\centering
		\begin{tabular*}{\linewidth}{@{} l @{\extracolsep{\fill}} rrrrr @{}}
			\hline
			\multicolumn{1}{@{} l}{\bfseries Dataset}
			& 
			\multicolumn{1}{r}{\bfseries RAND}
			&
			\multicolumn{1}{r}{\bfseries LC}
			& 
			\multicolumn{1}{r}{\bfseries M}
			& 
			\multicolumn{1}{r}{\bfseries EN}
			& 
			\multicolumn{1}{r @{}}{\bfseries LC-CL} \\
			\hline
			1\% & 39.83 & 39.83 & 39.83 & 39.83 & 39.83 \\
			10\% & 71.98 & 72.04 & 71.32 & 71.65 & 71.74 \\
			20\% & 73.83 & 76.01 & 75.61 & 75.87 & 75.43 \\
			30\% & 76.07 & 77.95 & 77.58 & 77.86 & 77.87 \\
			40\% & 78.42 & 79.45 & 79.08 & 79.69 & 79.50 \\
			50\% & 79.16 & 79.82 & 79.04 & 80.02 & 79.70 \\
			60\% & 80.00 & 80.39 & 80.01 & 80.29 & 79.91 \\
			70\% & 80.27 & \bfseries 81.50 & 80.55 & 81.09 & 80.77 \\
			80\% & 80.34 & 81.10 & 80.95 & 81.11 & 80.95 \\
			90\% & 80.93 & 81.40 & 81.07 & 81.02 & 81.02 \\
			100\% & 81.00 & 81.00 & 81.00 & 81.00 & 81.00 \\
			\hline
		\end{tabular*}
		\caption{EM measure of different \gls{al} strategies on SQuAD v1.1 dataset. The best performance result, specified in bold, belongs to the LC strategy on 70\% of the dataset, which outperforms the supervised method. RAND: random sampling stretegy. LC: least confidence. EN: entropy. M: margin sampling. LC-CL: our proposed clustering method.}
		\label{table:active_learning_em}
	\end{table}
\end{center}
% \vspace*{-0.05in}

\vspace*{-0.05in}
\begin{center}
	\begin{table}[H]
		\centering
		\begin{tabular*}{\linewidth}{@{} l @{\extracolsep{\fill}} rrrrr @{}}	
			\hline
			\multicolumn{1}{@{} l}{\bfseries Dataset}
			& 
			\multicolumn{1}{r}{\bfseries RAND}
			& 
			\multicolumn{1}{r}{\bfseries LC}
			& 
			\multicolumn{1}{r}{\bfseries M}
			&
			\multicolumn{1}{r}{\bfseries EN}
			&
			\multicolumn{1}{r @{}}{\bfseries LC-CL} \\
			\hline
			1\% & 50.64 & 50.64 & 50.64 & 50.64 & 50.64 \\
			10\% & 79.26 & 81.66 & 81.01 & 80.94 & 80.43 \\
			20\% & 82.81 & 84.83 & 84.31 & 84.67 & 84.40 \\
			30\% & 84.73 & 86.36 & 85.83 & 85.97 & 86.01 \\
			40\% & 86.51 & 87.50 & 86.79 & 87.44 & 87.21 \\
			50\% & 86.97 & 87.56 & 86.96 & 87.77 & 87.23 \\
			60\% & 87.69 & 88.05 & 87.75 & 87.89 & 87.88 \\
			70\% & 87.92 & \bfseries 88.60 & 88.00 & 88.39 & 88.12 \\
			80\% & 87.96 & 88.38 & 88.13 & 88.23 & 88.20 \\
			90\% & 88.12 & 88.56 & 88.25 & 88.27 & 88.24 \\
			100\% & 88.34 & 88.34 & 88.34 & 88.34 & 88.34 \\
			\hline
		\end{tabular*}
		\caption{F1-score measure of different \gls{al} strategies on SQuAD v1.1 dataset. The LC strategy on 70\% of the dataset, specified in bold, performs better than the supervised method. RAND: random sampling stretegy. LC: least confidence. EN: entropy. M: margin sampling. LC-CL: our proposed clustering method.}
		\label{table:active_learning_f1}
	\end{table}
\end{center}
\vspace*{-0.05in}

As it is shown in Table \ref{table:active_learning_f1}, using the least confidence strategy and only 20\% and 40\% of the training dataset, the model can respectively achieve 96.02\% and 99.04\% F1-score of the case in which we employ the supervised method and the whole dataset. As it can be seen in Tables \ref{table:active_learning_em} and \ref{table:active_learning_f1}, using 70\% of the training dataset and the least confidence strategy, the model can even outperform the supervised method by 0.50 and 0.26 in terms of the EM measure and F1-score, respectively. We think this is because \gls{al} is effectively using more informative samples for training and ignoring some noisy data. To the best of our knowledge, our work is the first application of \gls{al} to the \gls{qa} task.

\vspace*{-0.05in}
\subsection{Joint Application of Knowledge Distillation and Active Learning}
\vspace*{-0.05in}

In this section, to examine the joint application of \gls{kd} and \gls{al} to a single model, at first, 40\% of the training dataset was selected by the least confidence sampling method. Then, BERT\textsubscript{LARGE}, as the teacher model, was fine-tuned on this training set. Next, QANet was trained on the same dataset while its knowledge was being distilled using the teacher model. 

\vspace*{-0.05in}
\begin{center}
	\begin{table}[H]
		\centering
		\begin{tabular*}{\linewidth}{@{} l @{\extracolsep{\fill}} rrr @{}}
			\hline
			\multicolumn{1}{@{} l}{\bfseries Model}
			&
			\multicolumn{1}{r}{\bfseries Dataset}
			&
			\multicolumn{1}{r}{\bfseries F1}
			&
			\multicolumn{1}{r @{}}{\bfseries EM} \\
			\hline
			QANet (Base) \cite{yu2018qanet} & 100\% & 80.09 & 71.19 \\
			\hline
			BERT\textsubscript{LARGE} (Teacher) \cite{devlin-etal-2019-bert} & 40\% & 89.57 & 82.36 \\
			QANet (Base) \cite{yu2018qanet} & 40\% & 74.77 & 63.68 \\
			QANet+KD (Proposed) & 40\% & 78.83 & 68.31 \\
			QANet+KD+Linear (Proposed) & 40\% & \bfseries 79.51 & \bfseries 69.92 \\
			QANet+KD+Cubic (Proposed) & 40\% & 79.10 & 69.47 \\ 
			\hline
		\end{tabular*}
		\caption{Applying both \gls{kd} and \gls{al} on a single model trained on SQuAD v1.1 dataset. The results in bold show that our proposed model trained on 40\% of the dataset performs almost the same as the base model trained on 100\% of the dataset.}
		\label{table:both_al_kd}
	\end{table}
\end{center}
\vspace*{-0.05in}

The results of this experiment demonstrated in Table \ref{table:both_al_kd} show the QANet+KD+Linear model has outperformed the QANet (Base) model by 4.74 and 6.24 percentage points in F1 and EM, respectively, while trained on 40\% of the dataset. Besides, our model has achieved 99.20\% F1 and 98.20\% EM of the QANet (Base) model trained on 100\% of the dataset.

\vspace*{-0.05in}
\subsection{Robustness Against Adversarial Datasets}
\vspace*{-0.05in}

For analyzing the impact of \gls{kd} on the robustness of \gls{qa} models, QANet was trained and assessed on the Adversarial SQuAD dataset before and after applying \gls{kd}.

\vspace*{-0.05in}
\begin{center}
	\begin{table}[H]
		\centering
		\begin{tabular*}{\linewidth}{@{} l @{\extracolsep{\fill}} rrr @{}}
			\hline
			\multicolumn{1}{@{} l}{\bfseries Model}
			&
			\multicolumn{1}{r}{\bfseries Dataset}
			&
			\multicolumn{1}{r}{\bfseries F1}
			&
			\multicolumn{1}{r @{}}{\bfseries EM} \\
			\hline
			BERT\textsubscript{LARGE} \cite{devlin-etal-2019-bert} & 100\% & 67.81 & 63.10 \\
			BERT\textsubscript{BASE} \cite{devlin-etal-2019-bert} & 100\% & 53.79 & 48.20 \\
			\hline 
			QANet (Base) \cite{yu2018qanet} & 100\% & 41.91 & 36.20 \\
			QANet+KD (Proposed) & 100\% & 44.07 & 37.90 \\
			QANet+KD+Linear (Proposed) & 100\% & 45.91 & 40.10 \\
			QANet+KD+Cubic (Proposed) & 100\% & \bfseries 45.84 & \bfseries 40.30 \\
			\hline
			QANet (Base) \cite{yu2018qanet} & 40\% & 37.29 & 31.00 \\
			QANet+KD (Proposed) & 40\% & 41.07 & 34.60 \\
			QANet+KD+Linear (Proposed) & 40\% & 41.02 & 34.30 \\
			QANet+KD+Cubic (Proposed) & 40\% & \bfseries 41.24 & \bfseries 35.00 \\
			\hline
		\end{tabular*}
		\caption{Performance of our proposed models trained on SQuAD v1.1 dataset and evaluated on AddSent adversarial dataset. The best results, in bold, belongs to our proposed models trained on 100\% and 40\% of the dataset, which demonstrates the substantial impact of \gls{kd} on the robustness of the models.}
		\label{table:performance_on_addsent_adversarial}
	\end{table}
\end{center}
\vspace*{-0.05in}

\vspace*{-0.05in}
\begin{center}
	\begin{table}[H]
		\centering
		\begin{tabular*}{\linewidth}{@{} l @{\extracolsep{\fill}} rrr @{}}
			\hline
			\multicolumn{1}{@{} l}{\bfseries Model}
			&
			\multicolumn{1}{r}{\bfseries Dataset}
			&
			\multicolumn{1}{r}{\bfseries F1}
			&
			\multicolumn{1}{r @{}}{\bfseries EM} \\
			\hline
			BERT\textsubscript{LARGE} \cite{devlin-etal-2019-bert} & 100\% & 76.92 & 71.70 \\
			BERT\textsubscript{BASE} \cite{devlin-etal-2019-bert} & 100\% & 64.80 & 58.00 \\
			\hline 
			QANet (Base) \cite{yu2018qanet} & 100\% & 50.74 & 43.50 \\
			QANet+KD (Proposed) & 100\% & 53.64 & 46.20 \\
			QANet+KD+Linear (Proposed) & 100\% & 55.68 & 49.10 \\
			QANet+KD+Cubic (Proposed) & 100\% & \bfseries 55.90 & \bfseries 49.40 \\
			\hline
			QANet (Base) \cite{yu2018qanet} & 40\% & 46.58 & 38.30 \\
			QANet+KD (Proposed) & 40\% & 50.50 & 42.40 \\
			QANet+KD+Linear (Proposed) & 40\% & 50.47 & 42.40 \\
			QANet+KD+Cubic (Proposed) & 40\% & \bfseries 50.72 & \bfseries 43.00 \\
			\hline
		\end{tabular*}
		\caption{Performance of our proposed models trained on SQuAD v1.1 dataset and evaluated on AddOneSent adversarial dataset. The best experiment results, in bold, of our models on this type of adversarial dataset exhibit the strong impact of \gls{kd} and \gls{al} least confidence strategy on the robustness of the models.}
		\label{table:performance_on_addonesent_adversarial}
	\end{table}
\end{center}
\vspace*{-0.05in}

The results of our experiments in Tables \ref{table:performance_on_addsent_adversarial} and \ref{table:performance_on_addonesent_adversarial} show that using \gls{kd} increases both F1-score and EM of the base model that is trained on 100\% of SQuAD v1.1 by around 4.00 and 5.00 percentage points when it is tested on the AddSent and AddOneSent datasets, respectively. We also evaluated the performance of the model on the adversarial datasets when the model is equipped with both \gls{kd} and \gls{al}. The QANet+KD+Cubic model has been trained on 40\% of SQuAD v1.1 and sampled by the least confidence strategy. On the AddSent adversarial dataset, our model has outperformed the QANet (Base) model, trained on 40\% of SQuAD v1.1, by around 4.00 percentage points in F1-score and EM. It has also achieved 98.40\% F1-score of the base model that is trained on 100\% of the training dataset. The evaluation of this model on the AddOneSent adversarial dataset shows that using only 40\% of SQuAD v1.1, it can almost reach the same F1-score and EM as the base model that is trained on the whole training dataset.

\section{Conclusions}
\label{conclusions}

In this paper, we have proposed a novel combination of an interpolated \gls{kd} and \gls{al} for QA systems, which is comparable to state-of-the-art models in this task. Our experiments showed that our model while having a fewer number of parameters, outperformed both DistilBERT and TinyBERT with four layers and was comparable with these models with six layers. With \gls{al} and using only 40\% of the training data, we achieved a 99.04\% F1-score of the supervised model trained on the whole dataset. Furthermore, we showed that our proposed approach further boosts the performance of \gls{qa} models by reducing both the complexity of the model and required training data at the same time. Additionally, by testing the model on adversarial datasets, we showed that using \gls{kd} can also increase the robustness of the model.

As our future work, one interesting direction would be to further improve the effectiveness of \gls{kd} by connecting the intermediate layers of the teacher and student models to transfer the knowledge between those layers. Recently, pre-trained models such as ALBERT \cite{lan2020albert}, XLNet \cite{yang2020xlnet}, and RoBERTa \cite{liu2019roberta} have been introduced that have managed to improve the performance in some downstream tasks. It is interesting to investigate the usage of these models as the teacher model to improve the performance in the \gls{qa} task, too. Also, it may be beneficial if a combination of multiple teacher models would be used as an ensemble model.

\vskip 0.2in
\bibliography{paper}
\bibliographystyle{theapa}

\end{document}

%% file: assets/ir_based_qa_systems_architecture.tex
\begin{tikzpicture}[line cap=rect]

	\node[
    % draw, 
		circle, 
		minimum width=0.1cm,
		minimum height=0.1cm,
	] (origin_left) at (0, 0) {};  
	
  \node[
    shape=document,
    % double copy shadow={
    %   shadow xshift=-0ex,
    %   shadow yshift=-0ex
    % },
    draw,
    fill=white,
    text width=5em,
		text centered,
    minimum height=3em,
    right=0.1cm of origin_left
  ] (question) {Question};

	\node[draw,
		rectangle,
		% thick,
		fill=black!5,
		minimum height=3em,
		minimum width=5em,
		text width=5em,
		text centered,		
		rounded corners,
    right=of question
	] (question_processing) {Question Processing};	  

	\node[draw,
    rectangle,
		% thick,
		fill=black!5,		
		minimum height=3em,
		minimum width=5em,
		text width=5em,
		text centered,
    rounded corners,
		right=of question_processing
		% below right=0em and 0em of model
	] (document_retrieval) {Document Retrieval};	

  \node[
    shape=document,
    double copy shadow={
      shadow xshift=-0.8ex,
      shadow yshift=0.8ex
    },
    draw,
    fill=white,
    text width=5em,
		text centered,
    minimum height=3em,
    right=1cm of document_retrieval
  ] (relevant_documents) {Relevant Documents};  

	\node[draw,
    rectangle,
		% thick,
		fill=black!5,		
		minimum height=3em,
		minimum width=5em,
		text width=5em,
		text centered,
    rounded corners,
		below=of relevant_documents
		% below right=0em and 0em of model
	] (passage_retrieval) {Passage Retrieval};		

  \node[
    shape=document,
    double copy shadow={
      shadow xshift=-0.8ex,
      shadow yshift=0.8ex
    },
    draw,
    fill=white,
    text width=5em,
		text centered,
    minimum height=3em,
    left=0.8cm of passage_retrieval
  ] (passages) {Passages};  	

	\node[draw,
    rectangle,
		% thick,
		fill=black!5,		
		minimum height=3em,
		minimum width=5em,
		text width=5em,
		text centered,
    rounded corners,
		below=of question_processing
		% below right=0em and 0em of model
	] (answer_extraction) {Answer Extraction};	
	
	\node[
    shape=document,
    % double copy shadow={
    %   shadow xshift=-0ex,
    %   shadow yshift=-0ex
    % },
    draw,
    fill=white,
    text width=5em,
		text centered,
    minimum height=3em,
		below=of question
  ] (answer) {Answer};	

	\node[draw,
		rectangle,
		% thick,
		% fill=black!5,
		minimum height=3em,
		minimum width=5em,
		text width=5em,
		text centered,		
		rounded corners,
    above=of question_processing
	] (indexing) {Indexing};	

  \node[
    shape=document,
    double copy shadow={
      shadow xshift=-0.8ex,
      shadow yshift=0.8ex
    },
    draw,
    fill=white,
    text width=5em,
		text centered,
    minimum height=3em,
    left=of indexing
  ] (document) {Document};  

	\node[
    % draw, 
		circle, 
		minimum width=0.1cm,
		minimum height=0.1cm,
    above=0.1cm of document
	] (origin_top) {};    

	\node[
    % draw, 
		circle, 
		minimum width=0.1cm,
		minimum height=0.1cm,
    right=0.1cm of relevant_documents
	] (origin_right) {};

	\draw[thick, ->, >=latex] (question_processing) -- node[midway, xshift=1.5em]{} (document_retrieval);
  \draw[thick, ->, >=latex] (question) -- node[midway, xshift=1.5em]{} (question_processing);
	\draw[shorten <= 0.0cm, shorten >= 0.25cm, thick, ->, >=latex] (document_retrieval) -- node[midway, xshift=1.5em]{} (relevant_documents);
	\draw[shorten <= 1pt, shorten >= 0pt, thick, ->, >=latex] (relevant_documents) -- node[midway, xshift=1.5em]{} (passage_retrieval);
	\draw[shorten <= 0pt, shorten >= 0pt, thick, ->, >=latex] (passage_retrieval) -- node[midway, xshift=1.5em]{} (passages);
  \draw[shorten <= 0.25cm, shorten >= 0cm, thick, ->, >=latex] (passages) -- node[midway, xshift=1.5em]{} (answer_extraction);
  \draw[shorten <= 0cm, shorten >= 0cm, thick, ->, >=latex] (answer_extraction) -- node[midway, xshift=1.5em]{} (answer);
  \draw[shorten <= 0cm, shorten >= 0cm, thick, ->, >=latex] (indexing) -- node[midway, xshift=1.5em]{} (question_processing);
	\draw[shorten <= 0cm, shorten >= 0cm, thick, ->, >=latex] (document) -- node[midway, xshift=1.5em]{} (indexing);
  
	% \draw[thick, ->, >=latex] (model.270) -- (criteria.90);
	% \draw[thick] (pool.234) -- (criteria.0);
	% \draw[thick, ->, >=latex] (criteria) -- (samples);
	% \draw[thick, ->, >=latex] (samples) -- (annotator);
	% \draw[thick, ->, >=latex] (labeled.0) -- node[midway, above]{} (model.180);
	% \draw[thick, ->, >=latex] (origin) -- node[midway, above]{} (labeled);
							
	% \draw[thick] (unlabeled.east) -- node[midway, above]{\begin{tabular}{c} sample \end{tabular}} (model.west);
	% \draw[thick,->,>=stealth] (model.east) -- (candidates.west);
	% \draw[thick,->,>=stealth] (candidates.south) -- node[midway,xshift=-2.3em]{annotate} (human.north);
	% \draw[thick,->,>=stealth] (human.west) -- (labeled.east);
	% \draw[thick,->,>=stealth] (labeled.north) -- node[midway,xshift=-1.5em]{train} (model.south);
						
\end{tikzpicture}

%% file: assets/rc_based_qa_systems_architecture.tex
\begin{tikzpicture}[line cap=rect]

	\node[
    % draw, 
		circle, 
		minimum width=0.1cm,
		minimum height=0.1cm,
	] (origin_left) at (0, 0) {};  
	
  \node[
    shape=document,
    % double copy shadow={
    %   shadow xshift=-0ex,
    %   shadow yshift=-0ex
    % },
    draw,
    fill=white,
    text width=5em,
		text centered,
    minimum height=3em,
    right=0.1cm of origin_left
  ] (question) {Question};

	\node[draw,
		rectangle,
		% thick,
		fill=black!5,
		minimum height=3em,
		minimum width=5em,
		text width=5em,
		text centered,		
		rounded corners,
    right=of question
	] (question_processing) {Question Processing};	  

	\node[draw,
    rectangle,
		% thick,
		fill=black!5,		
		minimum height=3em,
		minimum width=7em,
		text width=7em,
		text centered,
    rounded corners,
		right=of question_processing
	] (machine_reading_comprehension) {Machine Reading Comprehension};		

  \node[
    shape=document,
    % double copy shadow={
    %   shadow xshift=-0.8ex,
    %   shadow yshift=0.8ex
    % },
    draw,
    fill=white,
    text width=5em,
		text centered,
    minimum height=3em,
    above=1cm of machine_reading_comprehension
  ] (passage) {Passage};  	
	
	\node[
    shape=document,
    % double copy shadow={
    %   shadow xshift=-0ex,
    %   shadow yshift=-0ex
    % },
    draw,
    fill=white,
    text width=5em,
		text centered,
    minimum height=3em,
		right=of machine_reading_comprehension
  ] (answer) {Answer};	

	\node[
    % draw, 
		circle, 
		minimum width=0.1cm,
		minimum height=0.1cm,
    above=0.1cm of passage
	] (origin_top) {};    

	\node[
    % draw, 
		circle, 
		minimum width=0.1cm,
		minimum height=0.1cm,
    right=0.1cm of answer
	] (origin_right) {};

  \draw[thick, ->, >=latex] (question) -- node[midway, xshift=1.5em]{} (question_processing);
	\draw[thick, ->, >=latex] (question_processing) -- node[midway, xshift=1.5em]{} (machine_reading_comprehension);
	\draw[thick, ->, >=latex] (machine_reading_comprehension) -- node[midway, xshift=1.5em]{} (answer);
	\draw[shorten <= 5pt, thick, ->, >=latex] (passage) -- node[midway, xshift=1.5em]{} (machine_reading_comprehension);
				
\end{tikzpicture}

%% file: assets/approach_1_knowledge_distillation.tex
\begin{tikzpicture}[x=1cm, y=1cm]
		
	% CONTEXT
	\node[draw, 
		rectangle, 
		% fill=black!10,
		minimum width=9em,
		minimum height=1.5em,
		rounded corners
	] (context) at (0, 0) {Context};
	% \node[draw,
	% 	circle,
	% 	fill=white,
	% 	minimum size=1em
	% ] (context_circle1) at ($(context.center) + (0, 0)$) {};
			
	% QUESTION
	\node[draw,
		rectangle,
		% fill=black!10,
		minimum width=9em,
		minimum height=1.5em,
		rounded corners
	] (question) at (10em, 0) {Question};

	% ANSWER
	\node[draw,
		rectangle,
		% fill=black!10,
		minimum width=1.5em,
		minimum height=7em,
		rounded corners
	] (answer) at (16em, 15.5em) {\rotatebox{90}{Answer}};	
			
	% % ANSWER
	% \node[draw,
	% 	rectangle,
	% 	% fill=black!10,
	% 	minimum width=9em,
	% 	minimum height=1.5em,
	% 	rounded corners
	% ] (answer) at (24em, 0) {Answer};
			
	% BERT
	\node[draw,
		rectangle,
		fill=black!5,
		minimum width=9em,
		minimum height=4em,
		text width=5em,
		text centered,
		rounded corners
	] (teacher) at (0, 6em) {BERT (teacher)};
	
	% QANET
	\node[draw,
		rectangle,
		fill=black!5,
		minimum width=9em,
		minimum height=4em,
		text width=5em,
		text centered,
		rounded corners
	] (student) at (10em, 6em) {QANet (student)};
	
	\draw[thick, ->, >=latex] (context.90) to (teacher.270);
	\draw[thick, dashed, ->, >=latex]
	($(question.90) + (-1em, 0)$) -- 
	($(question.90) + (-1em, 1em)$) --
	($(context.90) + (0.5em, 1em)$) --
	($(teacher.270) + (0.5em, 0)$);
	
	\draw[thick, densely dotted, ->, >=latex]
	($(context.90) + (0.5em, 0)$) -- 
	($(context.90) + (0.5em, 0.5em)$) --
	($(question.90) + (-0.5em, 0.5em)$) --
	($(student.270) + (-0.5em, 0)$);
	\draw[thick, ->, >=latex] (question.90) to (student.270);
	
	% BERT SOFTMAX
	\node[anchor=south west,
		draw,
		rectangle,
		% fill=black!10,
		minimum width=4em,
		minimum height=1.5em,
		text centered,
		rounded corners
	] (bert_sm_1) at (-4.4em, 12em) {softmax};	
	
	\node[anchor=south west,
		draw,
		rectangle,
		% fill=black!10,
		minimum width=4em,
		minimum height=1.5em,
		text centered,
		rounded corners
	] (bert_sm_2) at (0.2em, 12em) {softmax};
	
	\draw[thick, ->, >=latex] 
	(teacher.138) -- 
	node[midway, xshift=-1em]{$\alpha^1$} 
	(bert_sm_1.270);
	\draw[thick, ->, >=latex] 
	(teacher.41) -- 
	node[midway,xshift=-1em]{$\alpha^2$} 
	(bert_sm_2.270);

	%BERT KL
	\node[draw,
		rectangle,
		% fill=black!10,
		minimum width=4em,
		minimum height=1.5em,
		text centered,
		rounded corners,
		above=4em of bert_sm_1
	] (bert_kl_1) {KL};	
	
	\node[draw,
		rectangle,
		% fill=black!10,
		minimum width=4em,
		minimum height=1.5em,
		text centered,
		rounded corners,
		above=4em of bert_sm_2
	] (bert_kl_2) {KL};
	
	% QANET SOFTMAX
	\node[anchor=south west,
		draw,
		rectangle,
		% fill=black!10,
		minimum width=4em,
		minimum height=1.5em,
		text centered,
		rounded corners
	] (qanet_sm_1) at (10em-4.4em, 12em) {softmax};	
		
	\node[anchor=south west,
		draw,
		rectangle,
		% fill=black!10,
		minimum width=4em,
		minimum height=1.5em,
		text centered,
		rounded corners
	] (qanet_sm_2) at (10em+0.2em, 12em) {softmax};
	
	\draw[thick, ->, >=latex] 
	(student.138) -- 
	node[midway, xshift=-1em]{$\beta^1$}
	(qanet_sm_1.270);
	\draw[thick, ->, >=latex] 
	(student.41) -- 
	node[midway,xshift=-1em]{$\beta^2$} 
	(qanet_sm_2.270);
	
	%QANET CE
	\node[draw,
		rectangle,
		% fill=black!10,
		minimum width=4em,
		minimum height=1.5em,
		text centered,
		rounded corners,
		above=4em of qanet_sm_1
	] (qanet_ce_1) {CE};	
	
	\node[draw,
		rectangle,
		% fill=black!10,
		minimum width=4em,
		minimum height=1.5em,
		text centered,
		rounded corners,
		above=4em of qanet_sm_2
	] (qanet_ce_2) {CE};	

	%-----------------------------------------------------------------------------

	\draw[thick, ->, >=latex] 
	(qanet_sm_1.90) -- 
	node[midway, xshift=-1em, yshift=0.5em]{$p^1$}
	(qanet_ce_1.270);
	\draw[thick, ->, >=latex] 
	(qanet_sm_2.90) -- 
	node[midway, xshift=-1em, yshift=0.5em]{$p^2$}
	(qanet_ce_2.270);

	\draw[thick, dashed, ->, >=latex] 
	($(qanet_sm_1.90) + (-0.5em, 0)$) --
	($(qanet_sm_1.90) + (-0.5em, 0.5em)$) -- 
	($(bert_sm_1.90) + (0.5em, 0.5em)$) -- 
	($(bert_kl_1.270) + (0.5em, 0)$);
	\draw[thick, densely dotted, ->, >=latex]
	($(qanet_sm_2.90) + (-0.5em, 0)$) --
	($(qanet_sm_2.90) + (-0.5em, 1em)$) --
	($(bert_sm_2.90) + (0.5em, 1em)$) -- 
	($(bert_kl_2.270) + (0.5em, 0)$);

	\draw[thick, ->, >=latex]
	(bert_sm_1.90) --
	node[midway, xshift=-1em, yshift=0.5em]{$q^1$}
	(bert_kl_1.270);
	\draw[thick, ->, >=latex]
	(bert_sm_2.90) --
	node[midway, xshift=-1em, yshift=0.5em]{$q^2$}
	(bert_kl_2.270);

	\draw[thick, ->, >=latex]
	(answer.211) --
	($(qanet_sm_1.90) + (0.5em, 1.5em)$) --
	node[midway, xshift=1em, yshift=-0.25em]{$y^1$}
	($(qanet_ce_1.270) + (0.5em, 0)$);

	\draw[thick, ->, >=latex]
	($(answer.211) + (0, 0.5em)$) --
	($(qanet_sm_2.90) + (0.5em, 2em)$) --
	node[midway, xshift=1em, yshift=-0.25em]{$y^2$}
	($(qanet_ce_2.270) + (0.5em, 0)$);

	%-----------------------------------------------------------------------------

	% CSOFT
	\node[draw,
		rectangle,
		% fill=black!10,
		minimum width=9em,
		minimum height=2em,
		text width=5em,
		text centered,
		rounded corners
	] (csoft) at (0, 22em) {$C_{soft}$};
	
	% CHARD
	\node[draw,
		rectangle,
		% fill=black!10,
		minimum width=9em,
		minimum height=2em,
		text width=5em,
		text centered,
		rounded corners
	] (chard) at (10em, 22em) {$C_{hard}$};

	\draw[thick, ->, >=latex] (bert_kl_1.90) -- (csoft.204);
	\draw[thick, ->, >=latex] (bert_kl_2.90) -- (csoft.336);

	\draw[thick, ->, >=latex] (qanet_ce_1.90) -- (chard.204);
	\draw[thick, ->, >=latex] (qanet_ce_2.90) -- (chard.336);

	%-----------------------------------------------------------------------------

	% LOSS
	\node[draw,
		rectangle,
		% fill=black!10,
		minimum width=9em,
		minimum height=2em,
		text width=5em,
		text centered,
		rounded corners
	] (loss) at (5em, 26em) {$L$};

	\draw[thick, ->, >=latex] 
	(csoft.90) --
	($(csoft.90) + (0, 1em)$) --
	($(loss.270) + (-0.5em, -1em)$) -- 
	($(loss.270) + (-0.5em, 0)$);

	\draw[thick, ->, >=latex] 
	(chard.90) --
	($(chard.90) + (0, 1em)$) --
	($(loss.270) + (0.5em, -1em)$) -- 
	($(loss.270) + (0.5em, 0)$);
					
\end{tikzpicture}

%% file: assets/qanet_model_architecture.tex
\begin{tikzpicture}[x=1cm, y=1cm]
		
	% CONTEXT
	\node[draw, 
		rectangle, 
		% fill=black!10,
		minimum width=9em,
		minimum height=1.5em,
		rounded corners
	] (context) at (0, 0) {Context};

	\node[draw, 
		rectangle, 
		% fill=black!10,
		minimum width=9em,
		minimum height=1.5em,
		rounded corners,
		above=1em of context
	] (context_embedding) {Embedding};	

	\node[draw, 
		rectangle, 
		fill=black!5,
		minimum width=9em,
		minimum height=1.5em,
		text width=7em,
		text centered,
		rounded corners,
		above=1em of context_embedding
	] (context_stacked_embedding_encoder_blocks) {Stacked Embedding Encoder Blocks};	

	\draw[thick, ->, >=latex] (context) -- (context_embedding);
	\draw[thick, ->, >=latex] (context_embedding) -- (context_stacked_embedding_encoder_blocks);
			
	% QUESTION
	\node[draw, 
		rectangle, 
		% fill=black!10,
		minimum width=9em,
		minimum height=1.5em,
		rounded corners
	] (question) at (10em, 0) {Question};	

	\node[draw, 
		rectangle, 
		% fill=black!10,
		minimum width=9em,
		minimum height=1.5em,
		rounded corners,
		above=1em of question
	] (question_embedding) {Embedding};	

	\node[draw, 
		rectangle, 
		fill=black!5,
		minimum width=9em,
		minimum height=1.5em,
		text width=7em,
		text centered,
		rounded corners,
		above=1em of question_embedding
	] (question_stacked_embedding_encoder_blocks) {Stacked Embedding Encoder Blocks};		

	\draw[thick, ->, >=latex] (question) -- (question_embedding);
	\draw[thick, ->, >=latex] (question_embedding) -- (question_stacked_embedding_encoder_blocks);

	% CONTEXT-QUERY ATTENTION
	\node[draw,
		rectangle,
		% fill=black!10,
		minimum width=12em,
		minimum height=2em,
		text width=10em,
		text centered,
		rounded corners
	] (context_query_attention) at (5em, 11em) {Context-Query Attention};

	\draw[thick, ->, >=latex] (context_stacked_embedding_encoder_blocks) -- (context_query_attention);
	\draw[thick, ->, >=latex] (question_stacked_embedding_encoder_blocks) -- (context_query_attention);

	\node[draw, 
		rectangle, 
		fill=black!5,
		minimum width=9em,
		minimum height=1.5em,
		text width=7em,
		text centered,
		rounded corners,
		above=1em of context_query_attention
	] (stacked_model_encoder_blocks_1) {Stacked Model Encoder Blocks};	

	\draw[thick, ->, >=latex] (context_query_attention) -- (stacked_model_encoder_blocks_1);

	\node[draw, 
		rectangle, 
		fill=black!5,
		minimum width=9em,
		minimum height=1.5em,
		text width=7em,
		text centered,
		rounded corners,
		above=1em of stacked_model_encoder_blocks_1
	] (stacked_model_encoder_blocks_2) {Stacked Model Encoder Blocks};	

	\draw[thick, ->, >=latex] (stacked_model_encoder_blocks_1) -- (stacked_model_encoder_blocks_2);	

	\node[draw, 
		rectangle, 
		fill=black!5,
		minimum width=9em,
		minimum height=1.5em,
		text width=7em,
		text centered,
		rounded corners,
		above=1em of stacked_model_encoder_blocks_2
	] (stacked_model_encoder_blocks_3) {Stacked Model Encoder Blocks};	

	\draw[thick, ->, >=latex] (stacked_model_encoder_blocks_2) -- (stacked_model_encoder_blocks_3);
	
	% CONCAT LEFT
	\node[draw, 
		rectangle, 
		% fill=black!10,
		minimum width=9em,
		minimum height=1.5em,
		rounded corners
	] (concat_left) at (0em, 25em) {Concat};

	\draw[thick, ->, >=latex] 
	(stacked_model_encoder_blocks_1.180) --
	($(stacked_model_encoder_blocks_1.180) + (-1em, 0)$) -- 
	($(concat_left.270) + (-0.5em, 0)$);

	\draw[thick, ->, >=latex] 
	(stacked_model_encoder_blocks_2.180) --
	($(stacked_model_encoder_blocks_2.180) + (-0.5em, 0)$) -- 
	($(concat_left.270) + (0, 0)$);	

	\node[draw, 
		rectangle, 
		% fill=black!10,
		minimum width=9em,
		minimum height=1.5em,
		rounded corners,
		above=1em of concat_left
	] (linear_left) {Linear};

	\draw[thick, ->, >=latex] (concat_left) -- (linear_left);
	
	\node[draw, 
		rectangle, 
		% fill=black!10,
		minimum width=9em,
		minimum height=1.5em,
		rounded corners,
		above=1em of linear_left
	] (softmax_left) {Softmax};
	
	\draw[thick, ->, >=latex] (linear_left) -- (softmax_left);
	
	\node[draw, 
		rectangle, 
		% fill=black!10,
		minimum width=9em,
		minimum height=1.5em,
		rounded corners,
		above=1em of softmax_left
	] (start_probability) {Start Probability};	

	\draw[thick, ->, >=latex] (softmax_left) -- (start_probability);

	% CONCAT RIGHT
	\node[draw, 
		rectangle, 
		% fill=black!10,
		minimum width=9em,
		minimum height=1.5em,
		rounded corners
	] (concat_right) at (10em, 25em) {Concat};	

	\draw[thick, ->, >=latex] 
	(stacked_model_encoder_blocks_1.0) --
	($(stacked_model_encoder_blocks_1.0) + (1em, 0)$) -- 
	($(concat_right.270) + (0.5em, 0)$);	

	\draw[thick, ->, >=latex] 
	(stacked_model_encoder_blocks_3.0) --
	($(stacked_model_encoder_blocks_3.0) + (0.5em, 0)$) -- 
	($(concat_right.270) + (0, 0)$);		

	\node[draw, 
		rectangle, 
		% fill=black!10,
		minimum width=9em,
		minimum height=1.5em,
		rounded corners,
		above=1em of concat_right
	] (linear_right) {Linear};

	\draw[thick, ->, >=latex] (concat_right) -- (linear_right);
	
	\node[draw, 
		rectangle, 
		% fill=black!10,
		minimum width=9em,
		minimum height=1.5em,
		rounded corners,
		above=1em of linear_right
	] (softmax_right) {Softmax};
	
	\draw[thick, ->, >=latex] (linear_right) -- (softmax_right);
	
	\node[draw, 
		rectangle, 
		% fill=black!10,
		minimum width=9em,
		minimum height=1.5em,
		rounded corners,
		above=1em of softmax_right
	] (end_probability) {End Probability};	
	
	\draw[thick, ->, >=latex] (softmax_right) -- (end_probability);	

	\node[
		% draw, 
		rectangle, 
		% fill=black!10,
		minimum width=9em,
		minimum height=1.5em,
		rounded corners,
	] (model) at (5em, 35em) {Model};		

	%-----------------------------------------------------------------------------

	% ONE ENCODER BLOCK

	\node[
		% draw, 
		circle, 
		% fill=black!10,
		minimum width=0.5em,
		minimum height=0.5em,
		text centered,
		rounded corners,
	] (encoder_input) at (20em, -2em) {};

	% OUTER FRAME
	\node[
		draw, 
		rectangle, 
		fill=black!5,
		minimum width=9em,
		minimum height=34.15em,
		text centered,
		rounded corners,
		above=0.75em of encoder_input
	] (outer_frame) {};	

	\node[draw, 
		rectangle, 
		fill=white,
		minimum width=7em,
		minimum height=1.5em,
		text width=6em,
		text centered,
		rounded corners,
		above=1.5em of encoder_input
	] (position_encoding) {Position Encoding};	

	\draw[thick, ->, >=latex] (encoder_input) -- (position_encoding);		

	% REPEAT FRAME
	\node[
		draw, 
		dashed,
		rectangle, 
		fill=black!5,
		minimum width=8.5em,
		minimum height=8.5em,
		text centered,
		rounded corners,
		above=0.25em of position_encoding
	] (repeat_frame) {};		

	\node[draw, 
		rectangle, 
		fill=white,
		minimum width=7em,
		minimum height=1.5em,
		text width=6em,
		text centered,
		rounded corners,
		above=1.5em of position_encoding
	] (layernorm_1) {Layernorm};

	\draw[thick, ->, >=latex] (position_encoding) -- (layernorm_1);	

	\node[draw, 
		rectangle, 
		fill=white,
		minimum width=7em,
		minimum height=1.5em,
		text width=6em,
		text centered,
		rounded corners,
		above=1em of layernorm_1
	] (conv_1) {Conv};

	\draw[thick, ->, >=latex] (layernorm_1) -- (conv_1);		

	\node[draw, 
		circle, 
		fill=white,
		minimum width=0.5em,
		minimum height=0.5em,
		text centered,
		rounded corners,
		above=1em of conv_1
	] (add_1) {\scriptsize{+}};

	\draw[thick, ->, >=latex] (conv_1) -- (add_1);	

	\node[
		% draw, 
		rectangle, 
		minimum width=3em,
		minimum height=1.5em,
		text centered,
		rounded corners,
		right=0em of add_1
	] (repeat) {\scriptsize{Repeat}};	

	\draw[thick, ->, >=latex] 
	($(layernorm_1.270) + (0, -0.9em)$) --
	($(layernorm_1.270) + (-4em, -0.9em)$) --
	($(add_1.180) + (-3.1em, 0)$) -- 
	(add_1.180);

	\node[
		% draw, 
		rectangle, 
		minimum width=1em,
		minimum height=1em,
		text width=1em,
		text centered,
		rounded corners,
		above=1.5em of add_1
	] (rep) {\scriptsize{...}};

	\draw[thick, ->, >=latex] (add_1) -- (rep);		

	\node[draw, 
		rectangle, 
		fill=white,
		minimum width=7em,
		minimum height=1.5em,
		text width=6em,
		text centered,
		rounded corners,
		above=1.5em of rep
	] (layernorm_2) {Layernorm};

	\draw[thick, ->, >=latex] (rep) -- (layernorm_2);		

	\node[draw, 
		rectangle, 
		fill=white,
		minimum width=7em,
		minimum height=1.5em,
		text width=6em,
		text centered,
		rounded corners,
		above=1em of layernorm_2
	] (self_attention) {Self-Attention};

	\draw[thick, ->, >=latex] (layernorm_2) -- (self_attention);	
	
	\node[draw, 
		circle, 
		fill=white,
		minimum width=0.5em,
		minimum height=0.5em,
		text centered,
		rounded corners,
		above=1em of self_attention
	] (add_2) {\scriptsize{+}};

	\draw[thick, ->, >=latex] (self_attention) -- (add_2);	

	\draw[thick, ->, >=latex] 
	($(layernorm_2.270) + (0, -0.9em)$) --
	($(layernorm_2.270) + (-4em, -0.9em)$) --
	($(add_2.180) + (-3.1em, 0)$) -- 
	(add_2.180);	

	\node[draw, 
		rectangle, 
		fill=white,
		minimum width=7em,
		minimum height=1.5em,
		text width=6em,
		text centered,
		rounded corners,
		above=1.5em of add_2
	] (layernorm_3) {Layernorm};

	\draw[thick, ->, >=latex] (add_2) -- (layernorm_3);		

	\node[draw, 
		rectangle, 
		fill=white,
		minimum width=7em,
		minimum height=1.5em,
		text width=6em,
		text centered,
		rounded corners,
		above=1em of layernorm_3
	] (feedforward_layer) {Feedforward Layer};

	\draw[thick, ->, >=latex] (layernorm_3) -- (feedforward_layer);	
	
	\node[draw, 
		circle, 
		fill=white,
		minimum width=0.5em,
		minimum height=0.5em,
		text centered,
		rounded corners,
		above=1em of feedforward_layer
	] (add_3) {\scriptsize{+}};

	\draw[thick, ->, >=latex] (feedforward_layer) -- (add_3);		

	\draw[thick, ->, >=latex] 
	($(layernorm_3.270) + (0, -0.9em)$) --
	($(layernorm_3.270) + (-4em, -0.9em)$) --
	($(add_3.180) + (-3.1em, 0)$) -- 
	(add_3.180);	

	\node[
		% draw, 
		circle, 
		fill=white,
		minimum width=0.5em,
		minimum height=0.5em,
		text centered,
		rounded corners,
		above=1.5em of add_3
	] (encoder_output) {};

	\draw[thick, ->, >=latex] (add_3) -- (encoder_output);		
	
	\node[
		% draw, 
		rectangle, 
		% fill=black!10,
		minimum width=6em,
		minimum height=1.5em,
		text width=6em,
		text centered,
		rounded corners,
	] (model) at (16em, 35em) {One Encoder Block};

\end{tikzpicture}